\pdfoutput=1 
\documentclass[preprint]{elsarticle}

\usepackage{pdflscape}
\usepackage{pdfpages}

\usepackage{graphicx}
\usepackage{amssymb}
\usepackage{subfig}
\usepackage{amsmath}
\usepackage{float}
\usepackage{url}
\usepackage{makecell}
\usepackage{mathrsfs}
\usepackage{blindtext}
\usepackage{wrapfig}
\usepackage{graphicx}

\newcommand{\citationneeded}[1][]{\textsuperscript{[citation needed]}}

\begin{document}

\begin{landscape}
\includepdf[pages=-, angle=90]{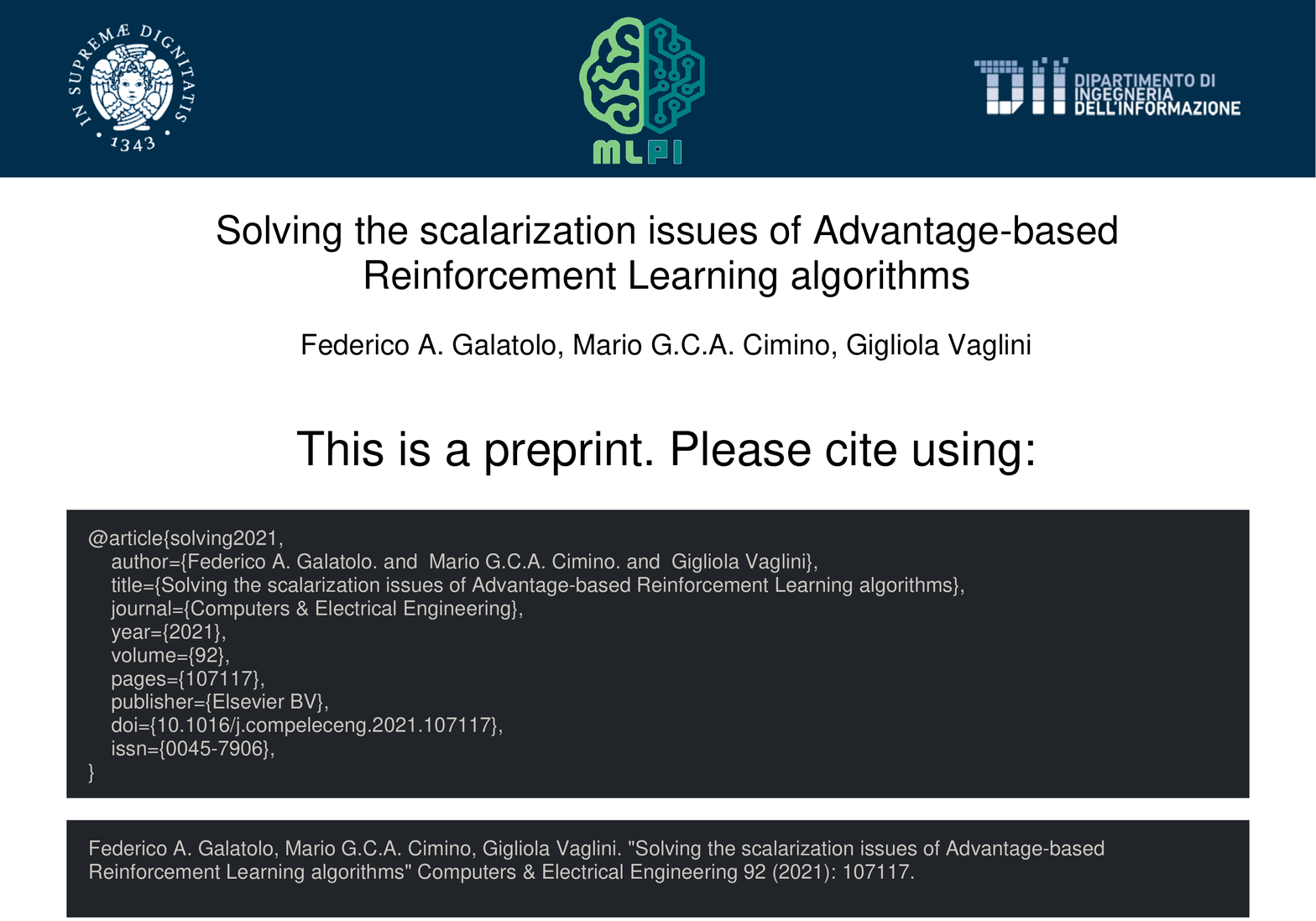}
\end{landscape}

\begin{frontmatter}
  \title{Solving the scalarization issues of Advantage-based Reinforcement Learning Algorithms}

  \cortext[cor1]{Corresponding author}
  \author{Federico A. Galatolo\corref{cor1}}
  \ead{federico.galatolo@ing.unipi.it}
  \author{Mario G.C.A. Cimino}
  \ead{mario.cimino@unipi.it}
  \author{Gigliola Vaglini}
  \ead{gigliola.vaglini@unipi.it}

  \address{Department of Information Engineering, University of Pisa, 56122 Pisa, Italy}

  \begin{abstract}
In this research, some of the issues that arise from the scalarization of the multi-objective optimization problem in the Advantage Actor Critic (A2C) reinforcement learning algorithm are investigated. The paper shows how a naive scalarization can lead to gradients overlapping. Furthermore, the possibility that the entropy regularization term can be a source of uncontrolled noise is discussed. With respect to the above issues, a technique to avoid gradient overlapping is proposed, while keeping the same loss formulation. Moreover, a method to avoid the uncontrolled noise, by sampling the actions from distributions with a desired minimum entropy, is investigated. Pilot experiments have been carried out to show how the proposed method speeds up the training. The proposed approach can be applied to any Advantage-based Reinforcement Learning algorithm.
  \end{abstract}
\end{frontmatter}

\section{Introduction and formal background}

\subsection{Introduction}

In last years, unprecedented results has been achieved in the Reinforcement Learning (RL) research field with the use of Artificial Neural Networks (ANNs). In essence, in an RL model an agent interacts with its environment and, upon observation of the consequences
of its actions, learns to adapt its own behaviour to rewards received. An agent behavior is modelled in terms of state-action relationships. 
The goal of the agent is to
learn a control strategy (i.e., a policy) maximizing the total reward. An important advancement in the field has been the possibility to operate with
high-dimensional state and action spaces via Deep Learning \cite{reinforcesurvey}.

More specifically, \textit{policy gradient} models  optimize the policy, represented as a parameterized function, via gradient-descent optimization.
An increasing interest of the research community has recently led to the paradigm shift of multi-objective reinforcement learning (MORL), in which learning
control policies are simultaneously optimized over several criteria
\cite{morl} \cite{gmorl}.

In RL \textit{Advantage learning} is used to estimate the advantage of performing a certain action. \cite{acktr}
Consequently, in the \textit{Actor-Critic (AC)} method a value function(which measures the expected reward) is learned in addition to the policy, in order to assist the policy update \cite{acsurvey}. This model is based on a ``Critic'', which estimates the value function, and an ``Actor'', which updates the policy distribution in the direction suggested by the ``Critic'' \cite{a3c}.

This research work focuses on some significant issues of the Advantage Actor Critic (A2C) algorithm, that arise from the scalarization of the multi-objective optimization problem. Firstly, it shows that a naive scalarization can lead to gradients overlapping. Secondly, it investigates the possibility that the entropy regularization term can inject uncontrolled noise. With respect to such issues, a technique to avoid gradient overlapping (called Non-Overlapping Gradient, NOG) is proposed, which keeps the same loss formulation. Moreover, a method to avoid the uncontrolled noise, by sampling the actions from distributions with a desired minimum entropy (called Target Entropy, TE), is investigated. Experimental results compare the A2C algorithm with the proposed combination of A2C with NOG and TE (A2C\textsubscript{NOG+TE}).

With regard to performance evaluation, we carried out the hyperparameters optimization for each scenario over the same task \cite{tpe}. Then using the best hyperparameters, we computed the confidence intervals over multiple runs.

As a relevant result, the combination of TE and NOG determines a decrease of the training time necessary to solve the problem. Specifically, the proposed technique achieves a larger speedup for increasing problem complexity.

The algorithmic design of the proposed approach is compliant with any Advantage-based Reinforcement Learning algorithm derived from A2C that share the same loss function components. 
The A2C\textsubscript{NOG+TE} algorithm has been developed, tested and publicly released on the Github platform, to foster its application on various research environments.

\subsection{Formal background}

An RL problem defines an \textit{environment} representing a task. The objective of an RL algorithm is to find an optimal \textit{policy} that an \text{agent} has to follow to solve the task.
The \textit{environment} can be represented as a Markov Decision Process (MDP). Denoting by $S$ the state space, and by $A$ the action space, it can be defined: (i) the \textit{state transition function} $f_s(s, a): \mathcal{S} \times \mathcal{A} \Rightarrow \mathcal{S}$; (ii)  the \text{reward function} $r(s, a): \mathcal{S} \times \mathcal{A} \Rightarrow \mathbb{R}$.

The objective of an RL algorithm is then to find a policy $\pi(s): \mathcal{S} \Rightarrow \mathcal{A}$ such that following its trajectories $\mathcal{T} = \{a_t = \pi(s_t), s_{t+1} = f_s(s_t, a_t) \;\; \forall t \}$ the cumulative sum of the rewards $\sum_{k=0}^{\infty} r(s_k, a_k)$ for any starting state $s_0$ is maximized.

Usually, the policy is stochastic: $\pi(s)$ is a function that, for each state $s \in \mathcal{S}$, returns the probability of each action $a \in \mathcal{A}$, i.e.,  $\pi(s): \mathcal{S} \Rightarrow \mathcal{A}\times (0, 1)$. By using $\pi(s, a)$ we assume that $a$ is the action \textit{sampled} from a categorical distribution with probabilities $\pi(s)$, and $\pi(s, a): \mathcal{S} \times \mathcal{A} \Rightarrow (0, 1)$ is the probability of the action $a$ in the distribution $\pi(s)$. Under such assumption, the objective is to maximize the \textit{expectation} of the cumulative sum of the rewards, i.e.,  $\mathbb{E}[\sum_{k=0}^{\infty} r(s_k, a_k)] = \sum_{k=0}^{\infty} r(s_k, a_k)\pi(s_k, a_k)$.

In the literature, if the \textit{policy} $\pi(s)$ is approximated using an ANN, the term Deep Reinforcement Learning is used. RL algorithms are divided into two major categories: \textit{off-policy} and \textit{on-policy} \cite{Tokic2012}. The \textit{off-policy} algorithms use stochastic techniques, for example $\epsilon-greedy$, to explore the state space. In the first phase, such algorithms perform random actions and accumulate the \textit{transactions} in a \textit{replay memory}. In the second phase, the \textit{off-policy} algorithms sample some \textit{transactions} from the \textit{replay memory}, and use them to train the \textit{policy}. In contrast, the \textit{on-policy} algorithms  explore the space by following the \textit{policy} and updating it via the current \textit{transactions} without a \textit{replay memory}.

In this paper we focus on the issues that arise in a family of \textit{on-policy} algorithms. 

\subsection{The Advantage Actor Critic (A2C) Algorithm}

The \textit{Advantage Actor Critic} (A2C) algorithm, proposed by \textit{OpenAI}, is the synchronous version of the \textit{Asynchronous Advantage Actor Critic} (A3C) algorithm, proposed by \textit{Google} \cite{a3c}. It has been shown that A2C has the same performance of A3C but with a lower implementation and execution complexity.

A2C is based on the REINFORCE algorithm \cite{acsurvey}. Let us define, for each time step $t$, the \textit{future discounted cumulative reward} $R_t = \sum_{i=0}^{\infty} \gamma^i r_{t+i}$. In the REINFORCE algorithm, each optimization step tends to maximize the expectation $E[R_t]$. Let us denote  $\theta_{\pi}$ the parameters of $\pi(s)$. The REINFORCE algorithm follows the optimization trajectory defined by $\Delta_{\theta_{\pi}} log(\pi(s, a | \theta_{\pi}))R_t$, which is an unbiased estimation of $\Delta_{\theta_{\pi}} E[R_t]$ \footnote{This is known as the \textit{log derivative trick}.}.

Usually, the quantity $log(\pi(s, a | \theta_{\pi}))R_t$ has an high variance, and the optimization trajectories defined by $\Delta_{\theta_{\pi}} log(\pi(s, a | \theta_{\pi}))R_t$ are very noisy. To overcome this issue a \textit{baseline} $b(t)$ is used to reduce the variance, and the gradient $\Delta_{\theta_{\pi}} log(\pi(s, a | \theta_{\pi}))(R_t - b(t))$ is computed. A classical \textit{baseline} can be the mean of $R_t$.

The contributions of A2C to REINFORCE are twofold: to use an ANN $V(s_t)$ approximating $R_t$ as the \textit{baseline} $b(t)$, and to use this ANN to bootstrap the $R_t$ computation in partially observed environmental trajectories. 

\begin{figure}
  \centering
  \includegraphics[width=\linewidth]{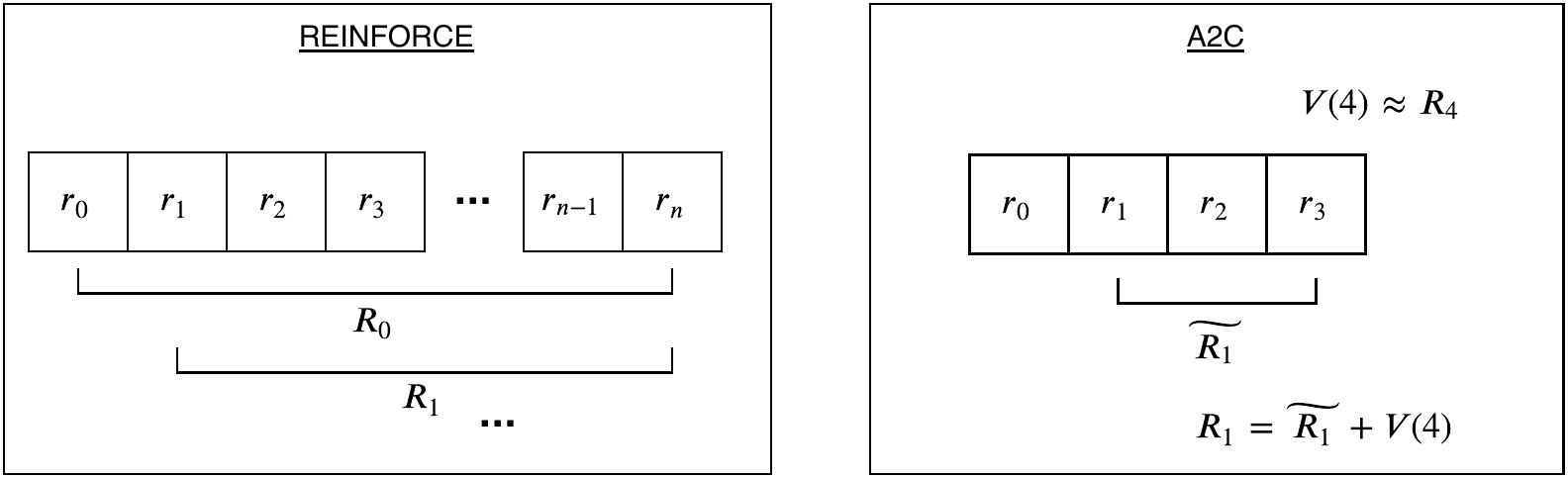}
  \caption[Caption for LOF]{$R_t$ computation in REINFORCE (left) and A2C\footnotemark} (right)
  \label{fig:reinforce_and_a2c}
\end{figure}

In REINFORCE $R_t$ can be computed after the end of the episode. In contrast, in A2C the $V(s_t)$ estimates $R_t$, and this value can be used to estimate the \textit{future discounted cumulative reward} before the end of the episode. Therefore, A2C performs an optimization step every $N$ steps, without waiting for the end of the episode. A visual representation of this difference is given in Figure \ref{fig:reinforce_and_a2c}. Here, each box represents the current reward $r_t$ whereas $R_t$ represents the future total discounted cumulative reward. In the case of A2C, the future total discounted cumulative reward is computed via the available cumulative reward $\tilde{R_t}$ and an estimation of $R_t$ of the last available state using $V$.\\

\footnotetext{For simplicity $R_t = \sum_{i=0}^{\infty} r_{t+i}$ is used in this example}
 
Overall, the remainder of this paper is structured as follows. Section 2 is devoted to the scalarization issues of the A2C algorithm. The proposed A2C\textsubscript{NOG+TE} algorithm is presented in Section 3. Experimental studies are covered by Section 4. Finally, Section 5 summarizes the major achievements and future work. 

\section{Scalarization issues of the A2C algorithm}

The A2C algorithm uses two ANNs to approximate the two functions $\pi(s | \theta_\pi)$ and $V(s | \theta_v)$. As previously stated, in A2C the environment is observed only for $N$ steps (instead of waiting for the episode termination). Given the partial \textit{state-action-reward} $(s_k, a_k, r_k) \forall k \in {t_s,\ldots,t_s+N}$ observation, the algorithm computes, for each $k$:
\begin{enumerate}
  \item $R_k$ using $V(s_{N+1})$ as bootstrap: $R_k = \sum_{i=k}^N\gamma^{i-k}r_{k} + \gamma^{N-k}V(s_{N+1})$;
  \item The policy gradient $\Delta_{pg} = \Delta_{\theta_{\pi}} log(\pi(s_k, a_k | \theta_{\pi}))(R_k - V(s_k))$;
  \item The $V(s | \theta_v)$ gradient $\Delta_{v} = \Delta_{\theta_v} (V(s_k | \theta_v) - R_k)^2$;
  \item The entropy gradient $\Delta_{h} = \Delta_{\theta_{\pi}} \sum_{i=0}^N log(\pi(s_i, a_i | \theta_{\pi}))\pi(s_i, a_i | \theta_{\pi})$.
\end{enumerate}

Subsequently, an optimization step is performed in the direction that maximizes both $\mathbb{E}[R_k]$ (direction $\Delta_{pg}$) and the entropy of $\pi(s_k)$ (direction $\Delta_h$), as well as minimizes the mean squared error of $V(s_k)$ (direction $-\Delta_v$). It is a multi-objective optimization problem, which in the A2C algorithm has been solved with a \textit{scalarization}. There are three different objectives, with some common parameters. Both the entropy and policy gradients share $\theta_{\pi}$.

Also $\pi(s)$ and $V(s)$ often have some common parameters, because usually a feature extraction is performed on the state $s$, and the features are  used as inputs for $\pi(s)$ and $V(s)$. Let us denote $C(s | \theta_C): \mathcal{S} \Rightarrow \mathcal{F}$ the feature extraction function, with $\theta_C$ its parameters, $f = C(s | \theta_C)$ the features. By substituting $\mathcal{S}$ with in $\mathcal{F}$ in the $\pi(s)$ and $V(s)$ domains \footnote{$\pi(f): \mathcal{F} \Rightarrow \mathcal{A}\times(0, 1)$, $\pi(f, a): \mathcal{F}\times\mathcal{A} \Rightarrow (0, 1)$ and $V(f): \mathcal{F} \Rightarrow R$}, then the computed gradients are:

\begin{align}
    \label{eq:a2c_delta_pg} \Delta_{pg} &= \Delta_{\theta_{\pi} + \theta_C} log(\pi(f_k, a_k | \theta_{\pi}, \theta_C))(R_k - V(f_k))\\
    \label{eq:a2c_delta_v} \Delta_{v}  &=  \Delta_{\theta_v + \theta_C} (V(f_k | \theta_v, \theta_C) - R_k)^2\\
    \label{eq:a2c_delta_h} \Delta_{h}  &= \Delta_{\theta_{\pi} + \theta_C} \sum_{i=0}^N log(\pi(s_i, a_i | \theta_{\pi}, \theta_C))\pi(s_i, a_i | \theta_{\pi}, \theta_C)
\end{align}
where $\Delta_{pg}$ is the policy gradient, $\Delta_v$ is the error gradient for the estimator net $V$, and $\Delta_{h}$ is a gradient of the entropy of the policy net. The notation $\Delta_{\theta_{\pi}+\theta_C}(\cdot)$ represents the gradient of the argument with respect to $\theta_{\pi}$ and $\theta_C$.

An optimization step is performed in the direction of the scalarized objective $-\Delta_{pg} + \beta\Delta_{v} - \alpha \Delta_{h}$, where $\alpha$ and $\beta$ are coefficients introduced to weight the strength of the entropy regularization term and of the $\Delta_v$ gradient, respectively. It is apparent that all the three objective functions share some parameters. Specifically, the gradient computed for the parameter $\theta_\pi$ contains the contributions of $\Delta_{pg}$ and $\Delta_{h}$. Furthermore, the gradient for the parameter $\theta_C$ contains the contributions of $\Delta_{pg}$, $\Delta_{v}$ and $\Delta_{h}$.

\begin{figure}[H]
  \centering
  \includegraphics[width=\linewidth]{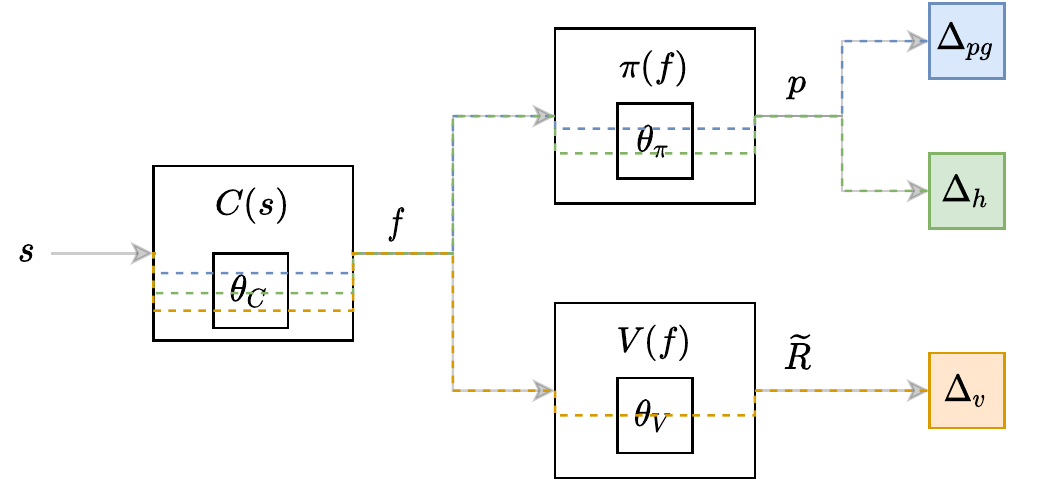}
  \caption{Backward computation in the A2C algorithm}
  \label{fig:a2c_gradients}
\end{figure}

A representation of the mutual dependency between gradients via related parameters is given in Figure \ref{fig:a2c_gradients}.

Each coloured box represents a different gradient contribution to the overall loss related to: the policy $\pi$, the entropy $h$, and the total  discounted cumulative reward estimator $V$.
Here, each big box represents a different Neural Network (NN), whereas the inner small box represents its parameters (i.e. a connection weights). In Figure, the input and output of each NN are also represented: $C(s)$ is fed by the state $s$ to extract the features $f$, whereas both the policy NN $\pi$ and the estimator NN $V$ take the features as an input, to provide the action probability vector $p$ and the future cumulative discounted reward estimate $\tilde{R}$, respectively.   
In particular, each gradient is represented with a different color, and a dashed colored arrow from the gradient to the inputs highlights the backward path and thus the influence of a gradient to a parameter optimization.
It is apparent that the sub-objectives are not independent, since they have common parameters. We call gradient overlapping this dependency among gradients. As a consequence, the policy and value function parameters can be pushed to sub-optimal regions.

Another issue that is considered in this research is the possibility that the entropy regularization term could generate noise in the network parameters. Indeed, it can be observed in Formula \ref{eq:a2c_delta_h} that the gradient $\Delta_h$ is not computed to reach a target entropy level, but just for increasing it.

In the next section the two issues are tackled considering also their reciprocal impact on the system performance.

\section{The Proposed A2C\textsubscript{NOG+TE} algorithm}

In this section a solution to avoid the gradient overlapping when using the A2C scalarized objective function is proposed. It is worth noting that to solve the gradient overlapping problem also allows to remove the weights coefficients of the scalarized objective function, thus reducing the hyperparameters search space, and then the optimization time. In the following, this approach will be referred to as the Non-Overlapping Gradient (NOG). Furthermore, an idea to solve the noise generated by the entropy regularization term is discussed. The idea is to maintain the entropy of the policy $\pi(f)$ above a target level without using any gradient. In the following, this approach will be referred to as the Target Entropy (TE). As an effect, this can further reduce the gradient overlapping phenomenon.

\subsection{Non-Overlapping-Gradients (NOG)}

The NOG technique consists in simplifying the backward computation flow represented in Fig. \ref{fig:a2c_gradients}, to remove the gradient overlapping on the feature extraction function $C(s)$, and to constrain the computation to the semantically appropriate functions. Specifically, the only gradient contributing to the feature extraction function $C(f)$ optimization is $\Delta_{pg}$. Similarly, the gradient $\Delta_h$ should contribute just to the policy function $\pi(f)$ optimization, as well as the gradient $\Delta_v$ should contribute just to the value function $V(f)$ optimization. According to such criterion, the new computed gradients are the following:

\begin{align}
    \label{eq:a2cnog_delta_pg} \Delta_{pg} &= \Delta_{\theta_{\pi}+ \theta_C} log(\pi(f_k, a_k | \theta_{\pi}, \theta_C))(R_k - V(f_k)) \\
    \label{eq:a2cnog_delta_v} \Delta_{v}  &= \Delta_{\theta_v} (V(f_k | \theta_v) - R_k)^2\\
    \label{eq:a2cnog_delta_h} \Delta_{h}  &= \Delta_{\theta_{\pi}} \sum_{i=0}^N log(\pi(s_i, a_i | \theta_{\pi}))\pi(s_i, a_i | \theta_{\pi})
\end{align}
where, with respect to Formulas \ref{eq:a2c_delta_pg}, \ref{eq:a2c_delta_v}, \ref{eq:a2c_delta_h}, $\Delta_{v}$ and $\Delta_h$ are computed respectively against $\theta_v$ and $\theta_{\pi}$ only.

\begin{figure}[H]
  \includegraphics[width=\linewidth]{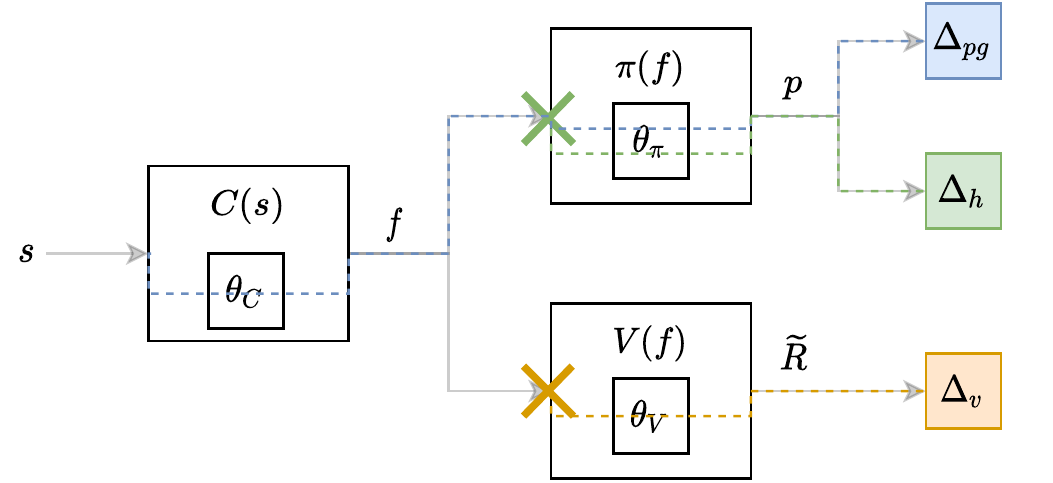}
  \caption{Backward computation in the A2C\textsubscript{NOG} algorithm}
  \label{fig:a2cnog_gradients}
\end{figure}

This way, the gradient overlapping is sensibly reduced, but not totally disappeared.
Specifically in this scenario the gradients $\Delta_{pg}$ and $\Delta_h$ still overlap via the parameters of the policy function $\theta_\pi$. A visual representation of the new gradient computation is given in Figure \ref{fig:a2cnog_gradients}, where a colored cross represents where the backward computation of the related gradient component stops.

Using the \textit{Non Overlapping Gradients} technique the new scalarized objective function is $-\Delta_{pg} + \Delta_{v} - \alpha \Delta_{h}$. Note that the parameter $\beta$ is not needed because $\Delta_{v}$ is totally independent.

\subsection{Target Entropy (TE)}

In the previous section, it has been highlighted that the gradient $\Delta_h$ is not computed to reach a target entropy level but just for increasing it. This can produce noise in the network parameters. In this section we propose a novel technique to maintain the entropy of the policy $\pi(f)$ above a target level without using any gradient. As a consequence, the gradient overlapping can be completely removed when using the TE technique in conjunction with \textit{NOG}.

Let us denote $p_a = \pi(f, a)$ the probabilities of each action $a \in \mathcal{A}$ given the features $f = C(s)$ of the state $s \in \mathcal{S}$, and  $p_{max}$ the highest probability. Let us observe that $\sum_i^N p_i = 1$. Let us define $\tilde{p}$ as:
\begin{align}
\tilde{p}_i=
\begin{cases}
p_i - \epsilon & i = max\\
p_i + \frac{\epsilon}{N-1} & i \neq max\\
\end{cases}
\label{eq:p_tilde}
\end{align}

The property $\sum_i^N \tilde{p}_i = 1$ is maintained \footnote{$\sum_i^N \tilde{p}_i = \sum_{i\neq max} \tilde{p}_i + p_{max} = \sum_{i \neq max} p_i - (N-1)\frac{\epsilon}{N-1} + p_{max} - \epsilon = \sum_{i \neq max} p_i + p_{max} -\epsilon + \epsilon = \sum_i^N p_i = 1$}, i.e., $\tilde{p}$ is still a valid categorical distribution. It is also important to notice that $H(\tilde{p}) < H(p)$.

Let us recall the definition of entropy $H(x) = -\sum_i^N log(x_i)x_i$, and let us focus on just one of the entropy components $log(x)x$. It can be easily compute the difference of one contribution in function of $\epsilon$  $\Delta h(x, \epsilon) = log(x)x - log(x+\epsilon)(x+\epsilon)$. Considering the overall entropy difference $\Delta H(p, \epsilon) = H(p) - H(\tilde{p}|\epsilon)$, it can be written in function of $\Delta h(p, \epsilon)$ contributions, as follows:

\begin{align*}
  \Delta H(p, \epsilon) = & \sum_i^N log(p_i)p_i - \sum_i^N log(\tilde{p_i})\tilde{p_i}\\
  \Delta H(p, \epsilon) = & log(p_0)p_0 + \cdots + log(p_n)p_n +\\
  &  -(log(p_0+\frac{\epsilon}{N-1})(p_0+\frac{\epsilon}{N-1}) + \cdots \\
  & + log(p_{n-1}+\frac{\epsilon}{N-1})(p_{n-1}+\frac{\epsilon}{N-1}) + log(p_{max}-\epsilon)(p_{max}-\epsilon) )\\
\end{align*}

Rearranging the terms, $\Delta H(p, \epsilon)$ can be rewritten as:

\begin{align*}
  \Delta H(p, \epsilon) = & (log(p_0)p_0 - log(p_0+\frac{\epsilon}{N-1})(p_0+\frac{\epsilon}{N-1})) + \\
  & \cdots \\
  & + (log(p_{n-1})p_{n-1} - log(p_{n-1}+\frac{\epsilon}{N-1})(p_{n-1}+\frac{\epsilon}{N-1}))\\
  & + log(p_{max}-\epsilon)(p_{max}-\epsilon)
\end{align*}

Expressing it in function of $\Delta_h$:

\begin{align*}
  \Delta H(p, \epsilon) = & \Delta_h(p_0, \frac{\epsilon}{N-1}) + \cdots  + \Delta_h(p_{n-1}, \frac{\epsilon}{N-1}) + \Delta_h(p_{max}, -\epsilon)
\end{align*}

Let us assume that $\epsilon$ is small and close to zero. The Taylor expansion of $\Delta_h(p, \epsilon)$ where $\epsilon = 0$ can be computed as follows:

\begin{align*}
  \frac{\partial}{\partial \epsilon}\Delta_h(p, \epsilon) = & -(\frac{1}{p+\epsilon}(p+\epsilon) + log(p+\epsilon)) = -log(p+\epsilon) - 1\\
  \Delta_h(p, \epsilon) |_{\epsilon \sim 0} \approx\;  & \Delta_h(p, 0) + \frac{\partial}{\partial \epsilon}\Delta_h(p, 0) \epsilon\\
  \Delta_h(p, \epsilon) |_{\epsilon \sim 0} \approx\;  & log(p)p - log(p)p + (-log(p)-1)\epsilon \approx -log(p)\epsilon - \epsilon\\
  \Delta_h(p, \epsilon) |_{\epsilon \sim 0} \approx\; & -\epsilon(log(p)+1)
\end{align*}

Finally, by substituting back the approximation of $\Delta_h(p, \epsilon)$ in $\Delta H(p, \epsilon)$, the following approximation can be derived:

\begin{align*}
  \Delta H(p, \epsilon) =\; & \Delta_h(p_0, \frac{\epsilon}{N-1}) + \cdots  + \Delta_h(p_{n-1}, \frac{\epsilon}{N-1}) + \Delta_h(p_{max}, -\epsilon)\\
  \Delta H(p, \epsilon) \approx\; & -\frac{\epsilon}{N-1}(log(p_0) + 1) \cdots -\frac{\epsilon}{N-1}(log(p_{n-1}) + 1) + \epsilon(log(p_{max})+1)\\
  \Delta H(p, \epsilon) \approx\; & -\frac{\epsilon}{N-1}(log(p_0)+log(p_1)+\cdots+log(p_{n-1}) - (N-1)(log(p_{max}) + 1) + (N-1))\\
  \Delta H(p, \epsilon) \approx\; & -\frac{\epsilon}{N-1}(\sum_i^{n-1} p_i - (N-1)log(p_{max}) - (N-1) + (N-1))\\
  \Delta H(p, \epsilon) \approx\; & -\epsilon (\frac{\sum_i^{n-1} p_i)}{N-1} - log(p_{max}))\\
  \Delta H(p, \epsilon) \approx\; & -\epsilon (AVG_{i\neq i_{max}}[p_i] - log(p_{max}))\\
\end{align*}

Using the above formula, $\epsilon$ can be computed as follows, in order to achieve a desired entropy $T_h$ of $p$:

\begin{align}
  \epsilon = -\frac{H(p) - T_h}{AVG_{i\neq i_{max}}[p_i] - log(p_{max})}
  \label{eq:epsilon}
\end{align}

\begin{figure}[H]
  \centering
  \includegraphics[width=\linewidth]{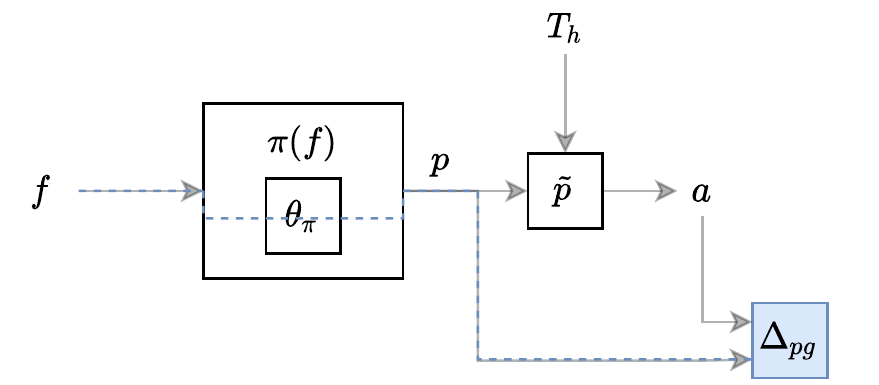}
  \caption{Backward computation in the A2C\textsubscript{TE} algorithm}
  \label{fig:a2cme_gradients}
\end{figure}

As a consequence, the action can be sampled from $\tilde{p} | \epsilon$ instead of $p$, i.e., to sample the action from a categorical distribution with an entropy higher than $T_h$. It is worth to notice that the action from the $\tilde{p} | \epsilon$ distribution can still be sampled  using the $\Delta_{pg}$ gradient computation represented in Figure \ref{fig:a2cme_gradients}. As a result, the technique allows to keep a certain \textit{exploration over exploitation} ratio, and at the same time it avoids raising  entropy.

Using the \textit{Target Entropy} technique, the new scalarized objective function is $-\Delta_{pg} + \beta\Delta_{v}$. It can be noted that there is no $\Delta_{h}$ term. Figure \ref{fig:a2cme_gradients} represents the resulting backward computation. Here, the focus is on the NN $\pi$, whose output $p$ is now transformed using the \textit{Target Entropy} according to \ref{eq:p_tilde} and \ref{eq:epsilon}. The resulting $\tilde{p}$ is used to sample an action $a$.
As a result, there is no more a contribution to the gradient related to $\Delta_h$. Then, the scalarization coefficients are not needed, because there are only two independent contributions to the gradient.

In the next section, the advantages of the NOG and TE techniques are experimentally evaluated.

\section{Experimental Studies}

In order to investigate the combined effect of the NOG and TE techniques, two different algorithms have been experimented:

\begin{enumerate}
    \item Classical A2C (A2C)
    \item A2C with Non-Overlapping-Gradients and Target Entropy (A2C\textsubscript{NOG+TE})
\end{enumerate}

For each training algorithm, first a hyperparameters optimization has been carried out. Subsequently, the best hyperparameters have been used to calculate the confidence interval, over 10 runs, of the training time needed to solve the problem.

To perform the experiments, three environments sufficiently complex to solve, which allow the hyperparameters optimization in a reasonable time, have been considered: \textit{EnergyMountainCar}, \textit{CartPole} and \textit{LunarLander}, all from OpenAI Gym \cite{gym}.

\subsection{Hyperparameters Optimization}

Table \ref{tab:hp} shows  the hyperparameters to optimize, for the considered algorithms.
In order to sample the hyperparameters to use for each run, it has been used the Tree-structured Parzen Estimator (TPE) \cite{tpe}, whereas to prune unpromising runs it has been used the Successive Halve Pruning (SHP) \cite{shp}. More precisely every 1000 steps the current reward EMA (Exponential Moving Average) is reported to the SHP pruner.

\begin{table}[!ht]
    \centering
  \begin{tabular}{||c c c c||} 
  \hline
  Name & Range & Sampling & Description \\ [0.5ex] 
  \hline\hline
  $\gamma$ & $[0.9, 0.99, 0.999]$ & Categorical & Discount factor \\  \hline
  $N$ & $[8, 16, 32, 64]$ & Categorical & Env. steps for training step \\  \hline
  lr &  $(10^{-5}, 10^{-2})$ & LogUniform & Learning rate \\  \hline
  mcn & $(0, 2)$ & Uniform & Max gradient clip norm\\ \hline
  $\alpha$ &  $(10^{-4}, 10^{-1})$ & LogUniform & $\Delta_h$ strength \\ \hline
  $\beta$ & $(0, 1)$ & Uniform & $\Delta_v$ strength \\ \hline
  $T_h$ &  $(0, 0.2)$ & Uniform & Target Entropy \\ 
  \hline
 \end{tabular}
  \caption{Hyperparameters to optimize.}
  \label{tab:hp}
\end{table}

Each run has been evaluated for 100 episodes, and the mean reward has been used as objective function (to maximize) for the hyperparameters optimization. All hyperparameters optimization has been run on an Intel Xeon with 40 cores.
It follows, for each environment, a brief description, the results of the hyperparameters optimization, and the performance evaluation for the two comparative algorithms.

\subsection{The EnergyMountainCar environment}

\begin{figure}[H]
  \centering
  \includegraphics[width=0.65\linewidth]{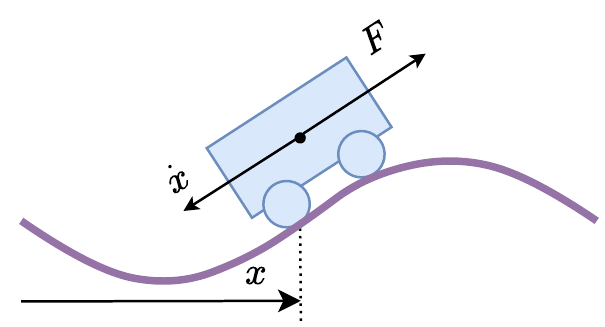}
  \caption{The EnergyMountainCar environment}
  \label{fig:energymountaincar}
\end{figure}

In \textit{EnergyMountainCar} a car drives up a hill which is steep with respect to its engine. Since the car is positioned in a valley, the agent must learn to drive back and forth to build up momentum.
Figure \ref{fig:energymountaincar} shows the environment and its control variables. 

Specifically, the state space has 2 components:  car's horizontal position ($x$) and horizontal speed ($\dot{x}$). Three different actions can be performed by the agent: no action, accelerate ($F$) to the left or to the right. The reward is computed as the car's total energy difference (potential and kinetic) of the last time step.

An episode finishes when the car reaches the top of the right hill. The goal is to spend less energy as possible.
The environment is considered solved by achieving a cumulative reward of 0.45 points.

\begin{figure}[H]
     \centering
     \subfloat[][A2C]{\includegraphics[width=0.8\linewidth]{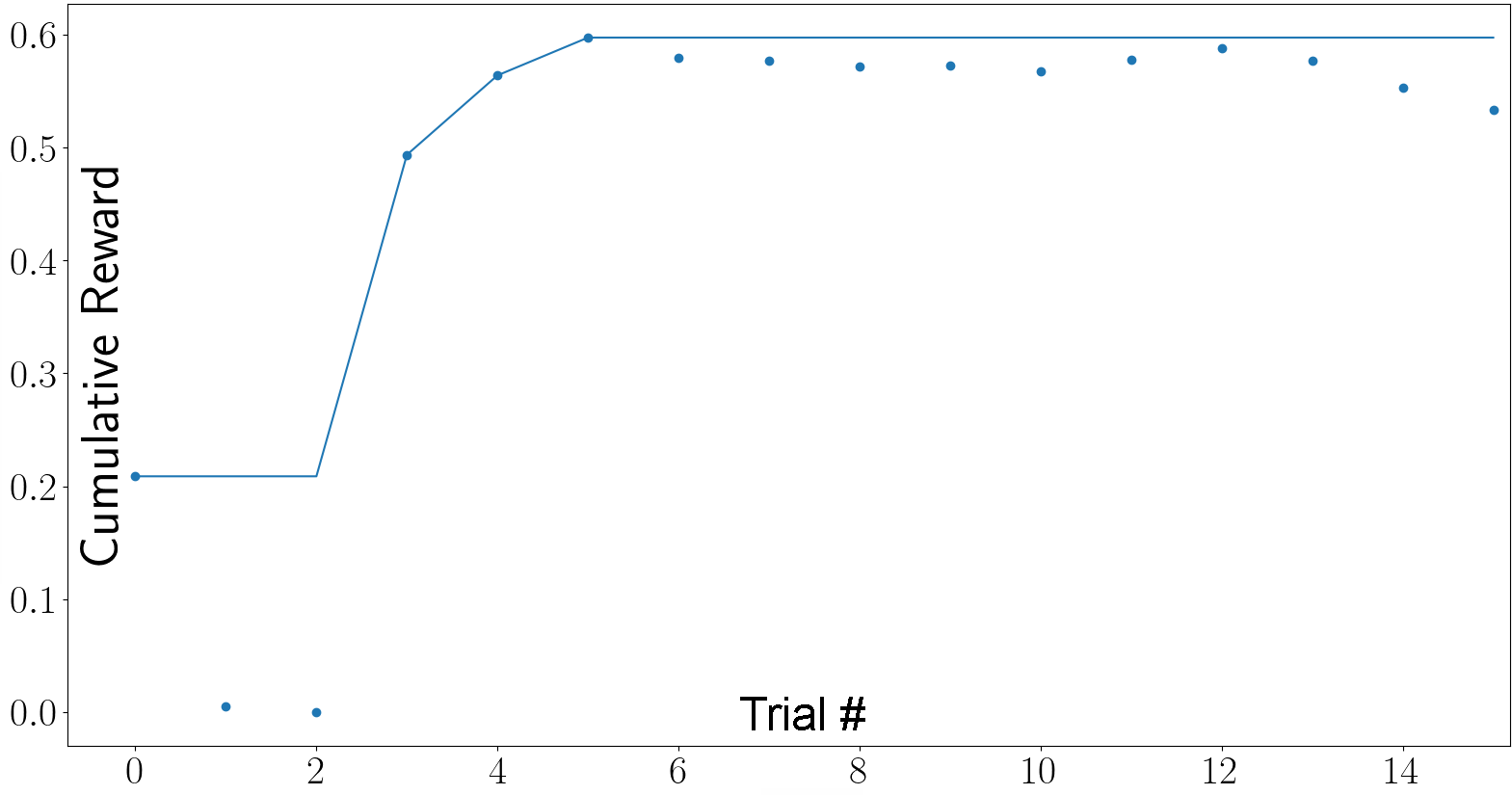}\label{fig:emc_A2C_time}}
     
     \subfloat[][A2C\textsubscript{NOG+TE}]{\includegraphics[width=0.8\linewidth]{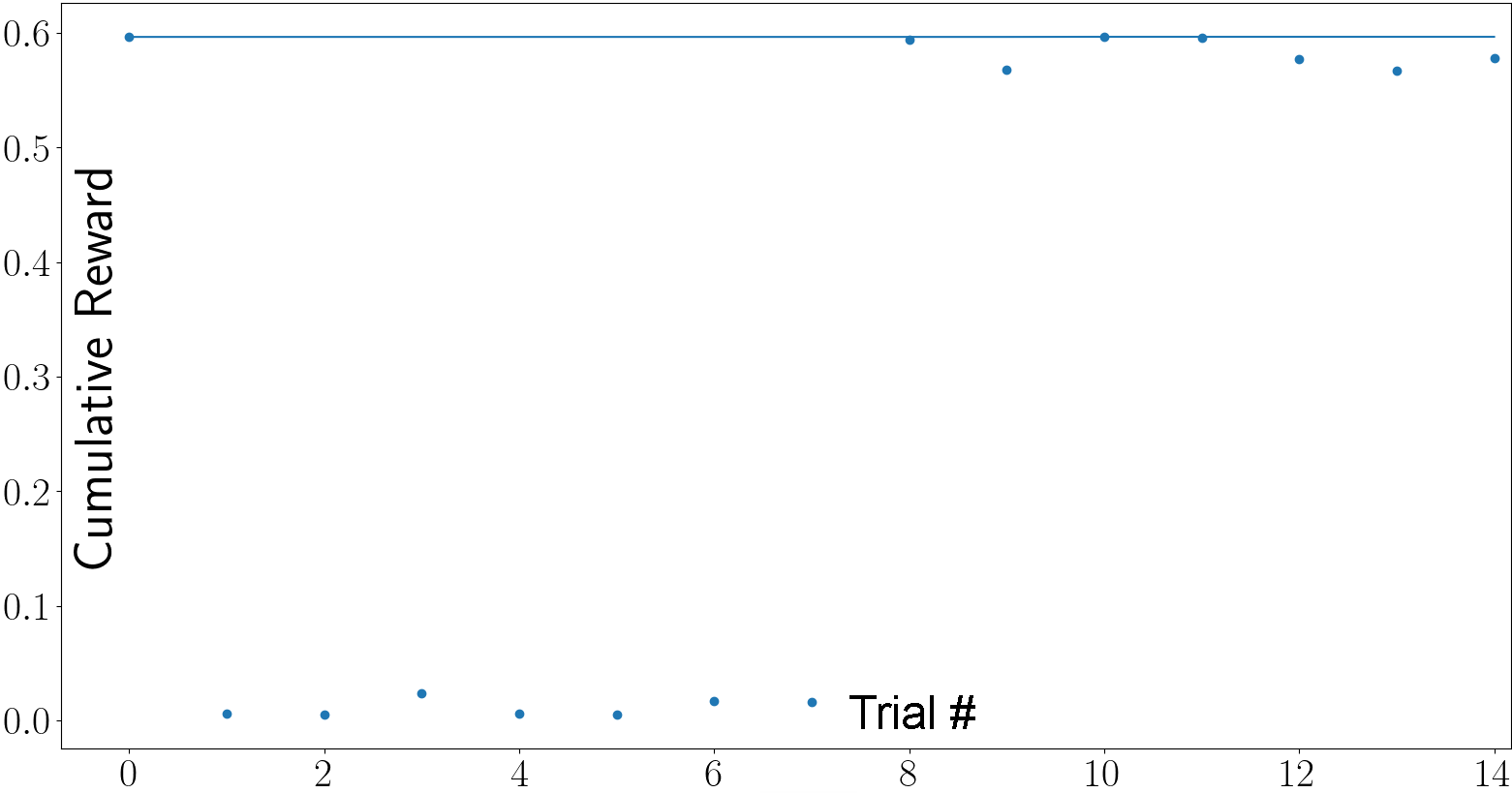}\label{fig:emc_A2Ctenog_time}}
     \caption{EnergyMountainCar: objective value of the hyperparameters optimization process over time, for the comparative algorithms. The solid line highlights the best values.}
     \label{fig:emc_time}
\end{figure}

Figure \ref{fig:emc_time} shows the objective value of the hyperparameters optimization process, against the number of trials sampled, for the comparative algorithm. The running best objective value is highlighted by a continuous line. In particular, it can be noted that the best objective value is immediately achieved by the A2C\textsubscript{NOG+TE}, whereas it is achieved at the fifth iteration by the A2C. 

\begin{figure}[H]
     \centering
     \subfloat[][A2C]{\includegraphics[width=0.9\linewidth]{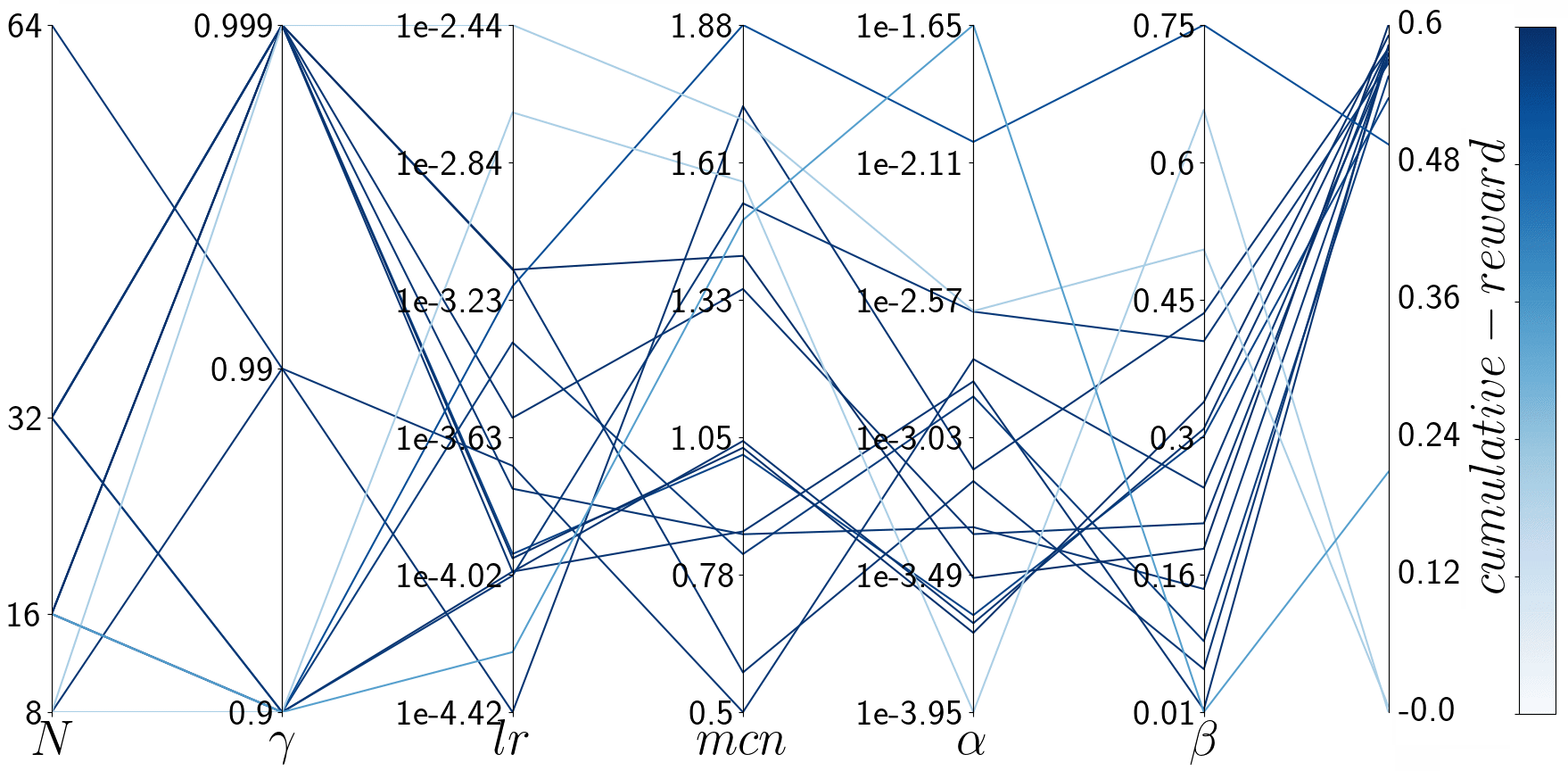}\label{fig:emc_A2C_parallel}}
     
     \subfloat[][A2C\textsubscript{NOG+TE}]{\includegraphics[width=0.9\linewidth]{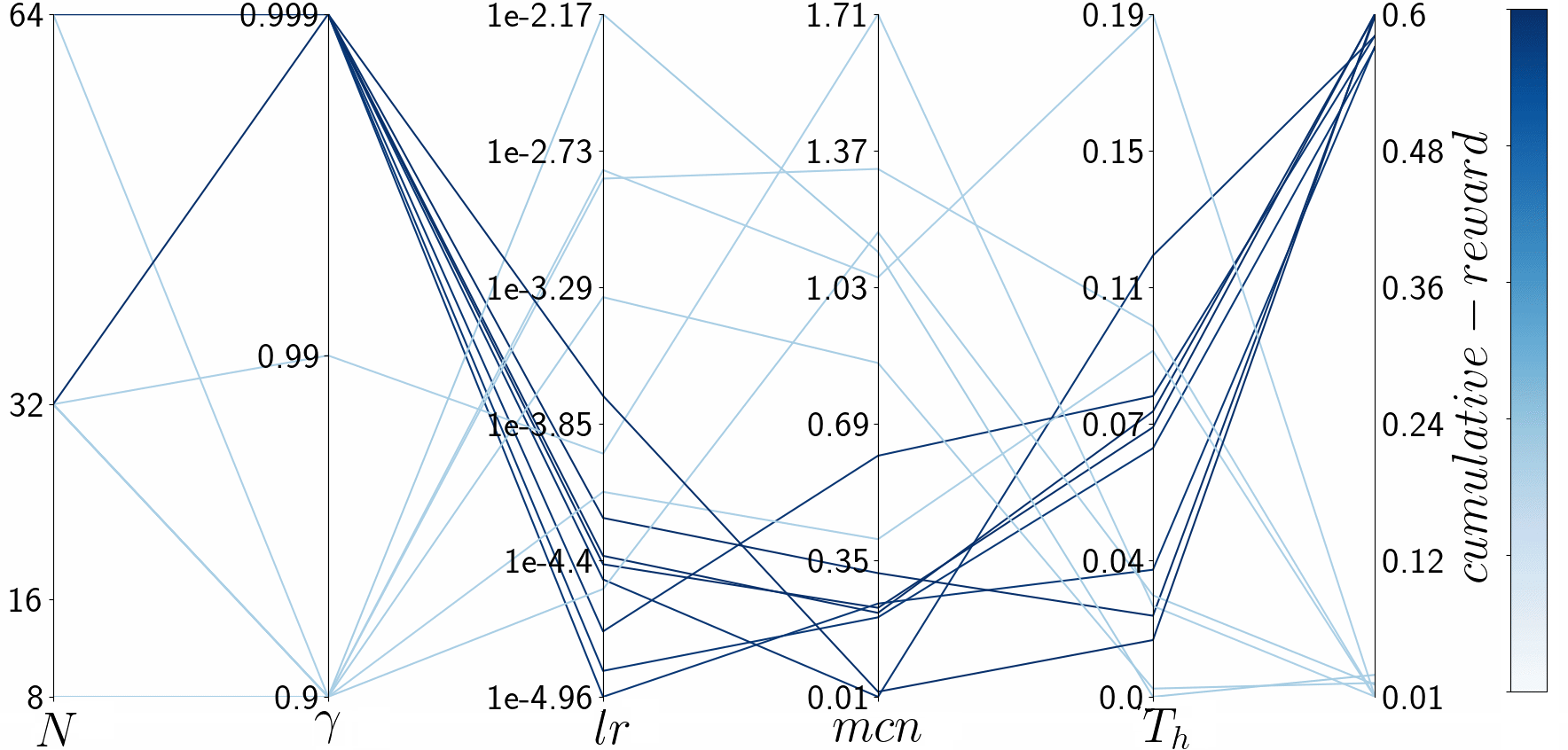}\label{fig:emc_A2Ctenog_parallel}}
     \caption{EnergyMountainCar: hyperparameters values optimization and related objective value, for the comparative algorithms.}
     \label{fig:emc_parallel}
\end{figure}

Figure \ref{fig:emc_parallel} represents the hyperparameters values optimization and the related objective value, for the two algorithms. Here, each line represents a trial, with its hyperparameters values represented on the vertical axes. According to the colorbar, the blue level of the line allows to distinguish the best solutions. Here, it can be observed that the hyperparameters values corresponding to the highest objective values are more scattered for the A2C.

For the sake of completeness, Table \ref{tab:emc_best_hp} shows the best hyperparameters value for each considered algorithm.

\begin{table}[!ht]
    \centering
  \begin{tabular}{||c c c ||} 
  \hline
  parameter & A2C & A2C\textsubscript{NOG+TE} \\ [0.5ex] 
  \hline\hline
  $\gamma$      & 0.999 & 0.999     \\
  \hline
  $N$           & 16        & 64        \\
  \hline
  lr            & 0.0007139 &  0.00003798 \\
  \hline
  max-clip-norm & 1.419     & 0.2302    \\
  \hline
  $\alpha$      & 0.0003160  & \\
  \hline
  $\beta$       & 0.1833 &  \\
  \hline
  $T_h$         &  & 0.0739 \\ 
  \hline
 \end{tabular}
  \caption{EnergyMountainCar: best hyperparameters found for each algorithm.}
  \label{tab:emc_best_hp}
\end{table}

After setting the best hyperparameters for each algorithm, the training process has been carried out 10 times for both algorithms.

\begin{figure}[H]
  \centering
  \includegraphics[width=.8\linewidth]{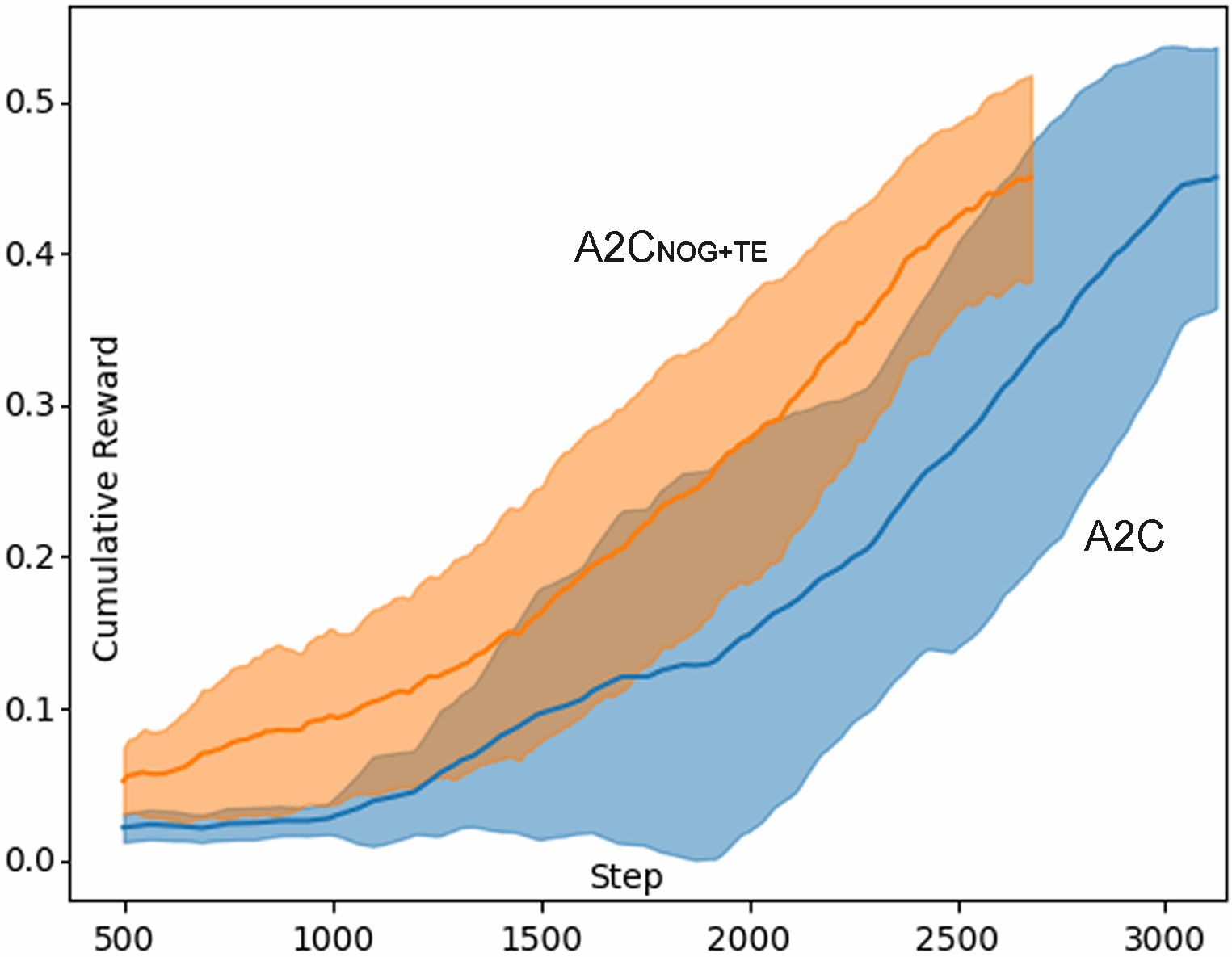}
  \caption{EnergyMountainCar: reward versus training step, for each algorithm.}
  \label{fig:energymountaincar_reward_vs_training}
\end{figure}

Figure \ref{fig:energymountaincar_reward_vs_training} shows the episode reward versus the training step for each algorithm, with its $95\%$ confidence interval. Precisely, the steps to solve the problem via the proposed  A2C\textsubscript{NOG+TE} algorithm and via the classical A2C are
$2511 \pm 378$ and $2702 \pm 433$, respectively. The proposed approach improves the time efficiency of the A2C, up to more than 1.08x of average speedup.

\subsection{The CartPole environment}

\begin{figure}[H]
  \centering
  \includegraphics[width=0.4\linewidth]{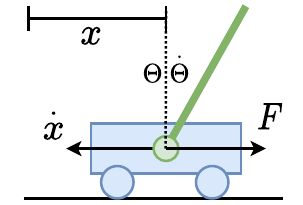}
  \caption{The CartPole environment}
  \label{fig:cartpole}
\end{figure}

\textit{CartPole}, also known as \textit{inverted pendulum}, is a pendulum with the center of mass above its pivot point. Figure \ref{fig:cartpole} shows the environment and its control variables. The pivot point is an axis of rotation mounted on a cart, limiting the pendulum to one degree of freedom, along which the cart can move horizontally. Any displacement from the vertical position causes a gravitation torque and a consequent fall, if not balanced by the cart movement.
The agent controls the cart in order to prevent the pendulum from falling, by applying a force $F$ of $\pm 1$. The state space is represented by 4 components: cart position ($x$), cart velocity ($\dot{x}$), pole angle ($\Theta$), and pole tip angular velocity ($\dot{\Theta}$). The action space is two-dimensional: moving left or right. A reward of +1 is provided for every timestep with the pole upright. An episode ends when the pole is more than 15 degrees from vertical, or when the cart moves more than 2.4 units from the start.
The goal is to keep the pole upright as much as possible.
The environment is considered solved by achieving a cumulative reward of 195 points.

\begin{figure}[H]
     \centering
     \subfloat[][A2C]{\includegraphics[width=0.8\linewidth]{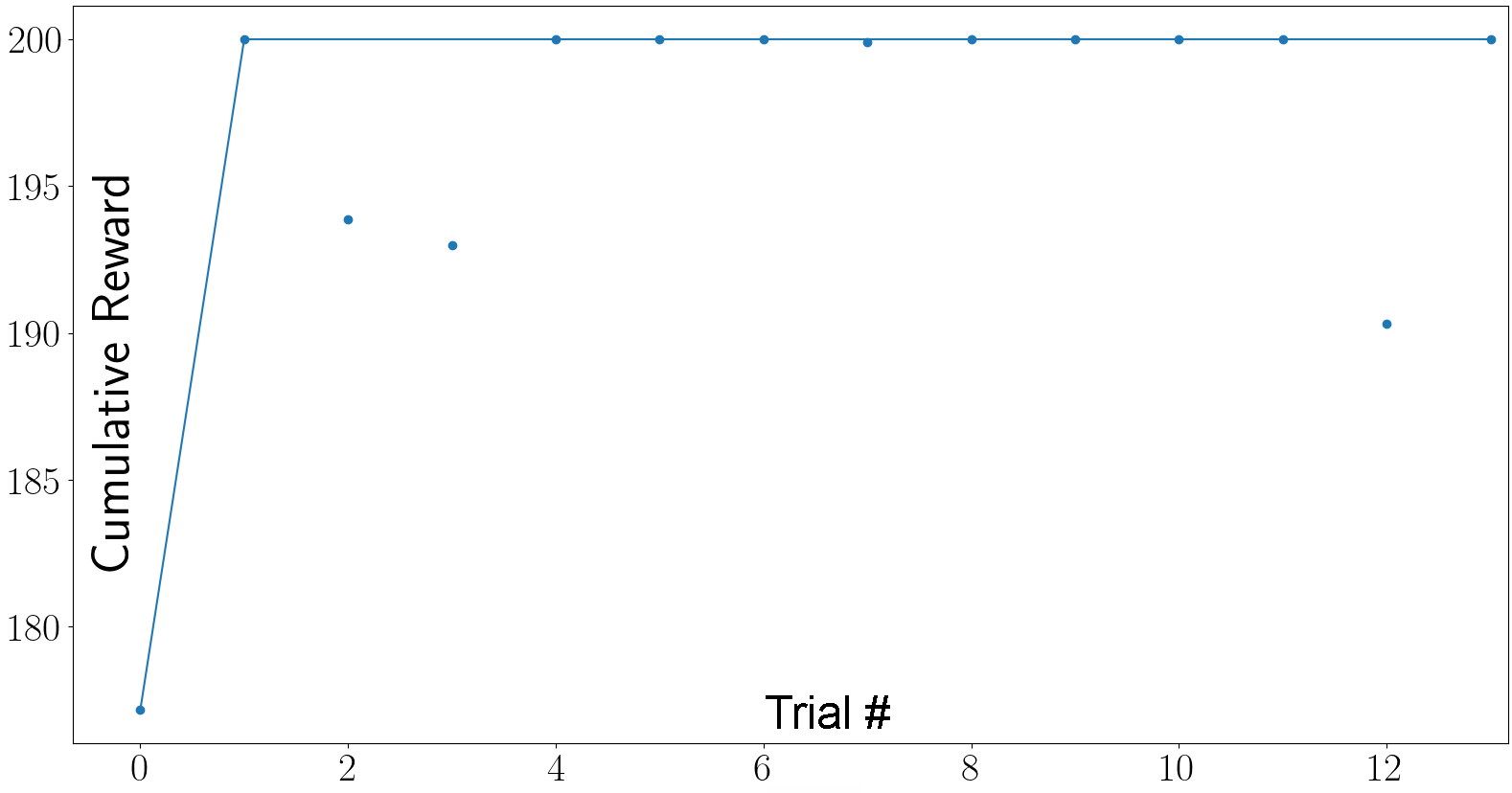}\label{fig:cp_A2C_time}} \\
     \subfloat[][A2C\textsubscript{NOG+TE}]{\includegraphics[width=0.8\linewidth]{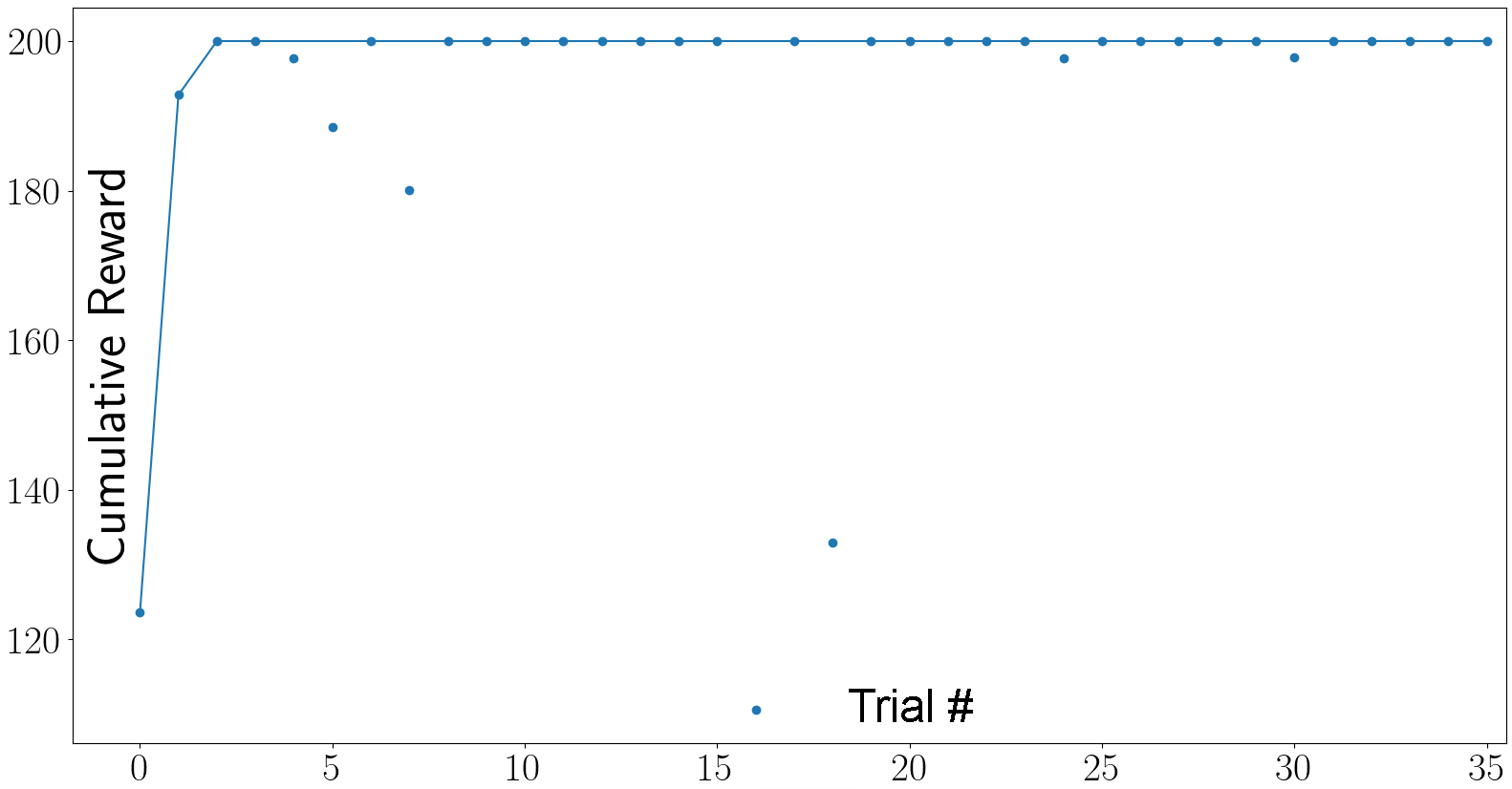}\label{fig:cp_A2Ctenog_time}}
     \caption{CartPole: objective value of the hyperparameters optimization process over time, for the comparative algorithms. The solid line highlights the best values.}
     \label{fig:cp_time}
\end{figure}

Figure \ref{fig:cp_time} shows the objective value of the hyperparameters optimization process, against the number of trials sampled, for the comparative algorithm. It is worth noting that, there is a lower number of trials in the hyperparameters optimization of A2C. Specifically, during the optimization, there are unpromising trials which are aborted during the training process by the pruning algorithm.
The figures clearly show that the A2C has been more affected by pruning with respect to the A2C\textsubscript{NOG+TE}. 

\begin{figure}[H]
     \centering
     \subfloat[][A2C]{\includegraphics[width=0.9\linewidth]{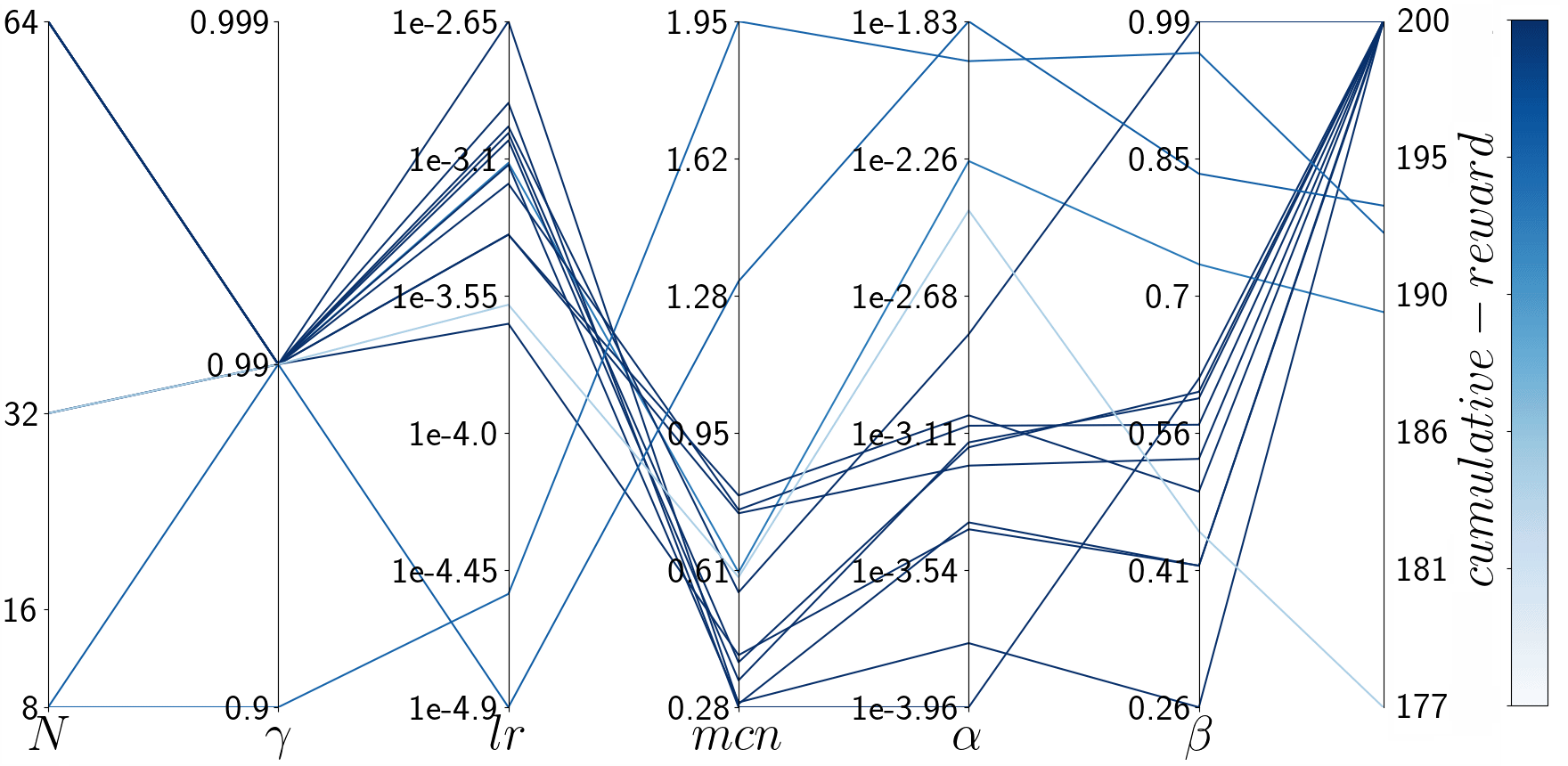}\label{fig:cp_A2C_parallel}}
     
     \subfloat[][A2C\textsubscript{NOG+TE}]{\includegraphics[width=0.9\linewidth]{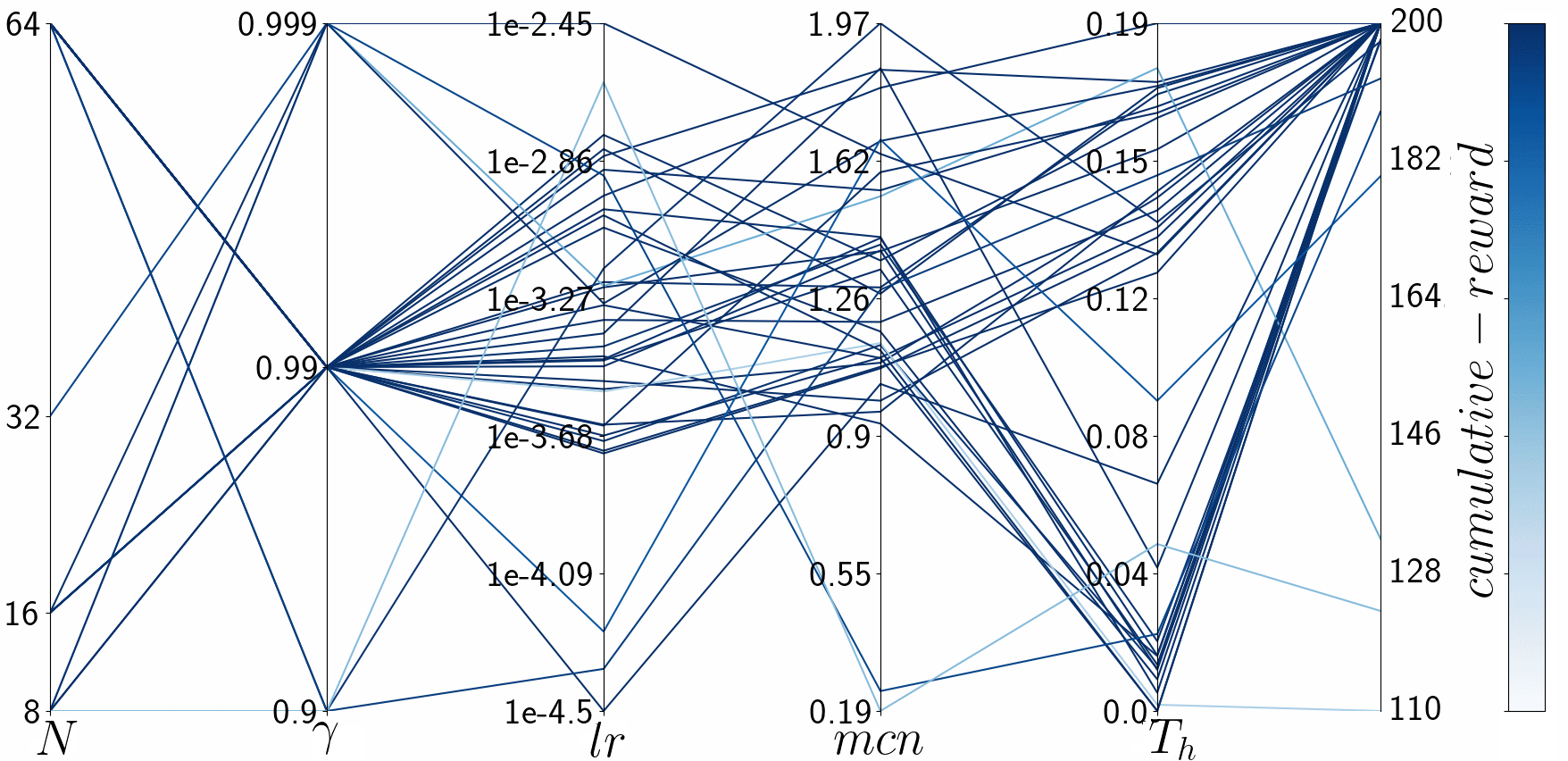}\label{fig:cp_A2Ctenog_parallel}}
     \caption{CartPole: hyperparameters values optimization and related objective value, for the comparative algorithms.}
     \label{fig:cp_parallel}
\end{figure}

Figure \ref{fig:cp_parallel} represents the hyperparameters values optimization and the related objective value, for the two algorithms. Here, each line represents a trial, with each hyperparameter value represented on the vertical axes. According to the colorbar, the blue level of the line allows to distinguish the best solutions. Here, it can be observed that for some hyperparameters values there is a higher density of good trials.

For the sake of completeness, Table \ref{tab:cp_best_hp} shows the best hyperparameters value for both algorithms.

\begin{table}[!ht]
    \centering
  \begin{tabular}{||c c c ||} 
  \hline
  parameter & A2C & A2C\textsubscript{NOG+TE} \\ [0.5ex] 
  \hline\hline
  $\gamma$      & 0.99 & 0.99     \\
  \hline
  $N$           & 64        & 64        \\
  \hline
  lr            & 0.0009591 &  0.001642 \\
  \hline
  max-clip-norm & 0.3898     & 1.3569    \\
  \hline
  $\alpha$      & 0.0006986  & \\
  \hline
  $\beta$       & 0.5996 &  \\
  \hline
  $T_h$         &  & 0.166 \\ 
  \hline
 \end{tabular}
  \caption{CartPole: best hyperparameters found for each algorithm.}
  \label{tab:cp_best_hp}
\end{table}

After setting the best hyperparameters for each algorithm, the training process has been carried out 10 times for each algorithm.

\begin{figure}[H]
  \centering
  \includegraphics[width=.8\linewidth]{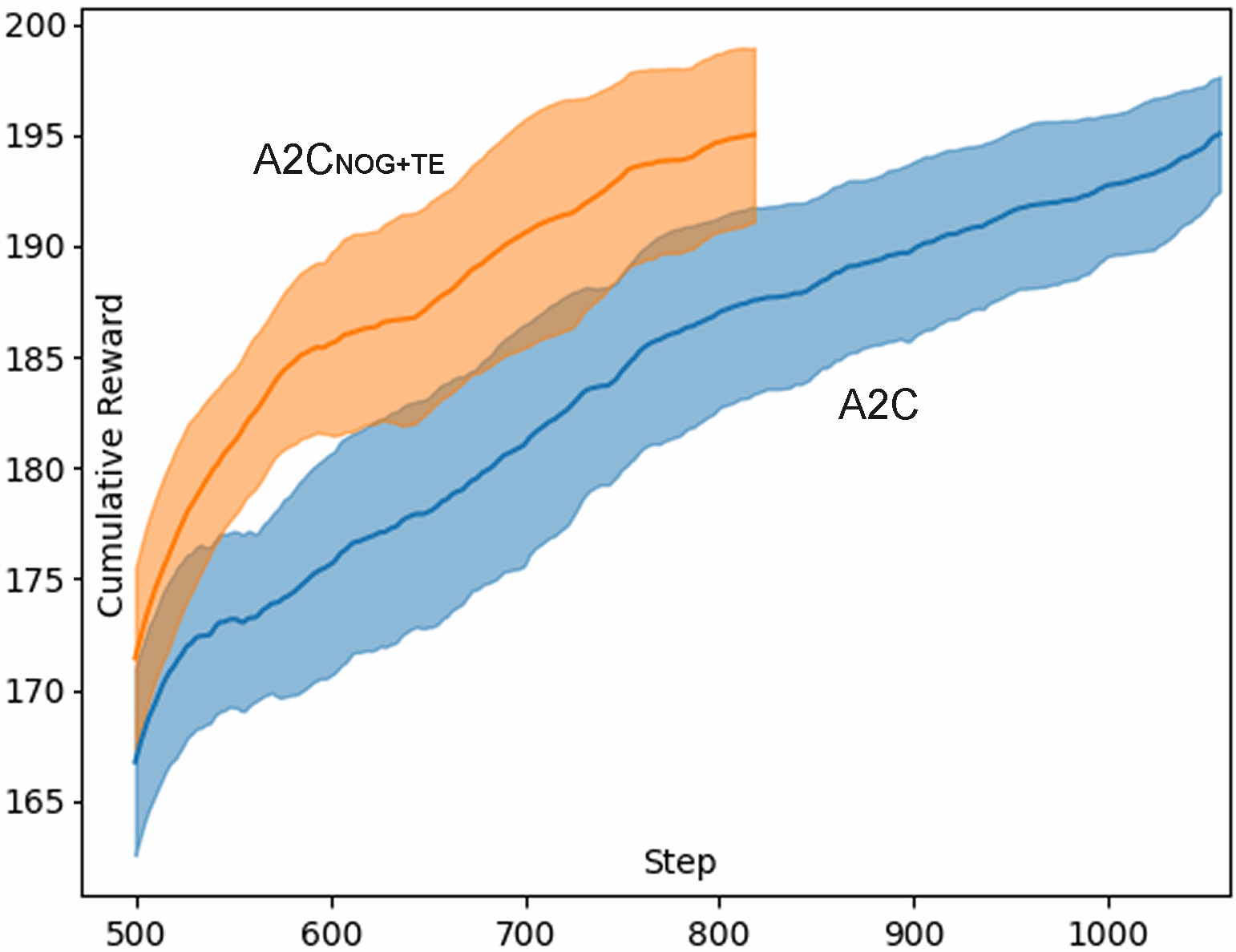}
  \caption{CartPole: reward versus training step, for each algorithm.}
  \label{fig:cartpole_reward_vs_training}
\end{figure}

Figure \ref{fig:cartpole_reward_vs_training} shows the episode reward versus the training step for each algorithm, with its $95\%$ confidence interval. Precisely, the steps to solve the problem via the proposed  A2C\textsubscript{NOG+TE} algorithm and via the classical A2C are
$848 \pm 197$ and $999 \pm 108$, respectively. 
The proposed approach sensibly improves the time efficiency of the A2C, up to more than 1.18x of average speedup.

\subsection{The LunarLander environment}

\begin{figure}[H]
  \centering
  \includegraphics[width=0.65\linewidth]{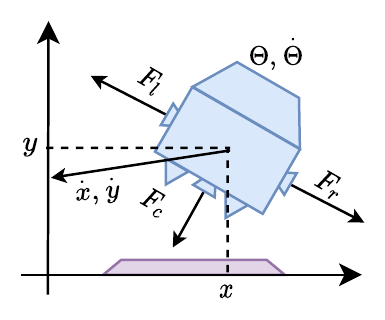}
  \caption{The LunarLander environment}
  \label{fig:lunarlander}
\end{figure}

\textit{LunarLander} is a control task, in which the agent controls the landing of a spacecraft. The spacecraft is initialized at the top of the environment, with a random velocity and angular momentum. Figure \ref{fig:lunarlander} shows the environment and its control variables. 

Specifically, the state space has 8 components: horizontal position and velocity ($x, \dot{x}$), vertical position and velocity ($y, \dot{y}$), angle ($\Theta$) and angular momentum ($\dot{\Theta}$), right and left leg state (that is, leg ground contact). Four different actions can be performed by the agent: no action, fire left engine ($F_l$), fire right engine ($F_r$) and fire main engine ($F_c$). The spacecraft has infinite fuel. The reward is computed as follows: -0.3 points for each frame with the main engine on, +100 points for a successful landing, -100 points for crashing, +10 points for each leg making contact with the ground, and a value ranging from 100 to 140 evaluating the spacecraft trajectory to the pad. An episode finishes when the spacecraft lands or crashes. The goal is to land the spacecraft using as less fuel as possible. The environment is considered solved by achieving a cumulative reward of 200 points.

\begin{figure}[H]
     \centering
     \subfloat[][A2C]{\includegraphics[width=1.0\linewidth]{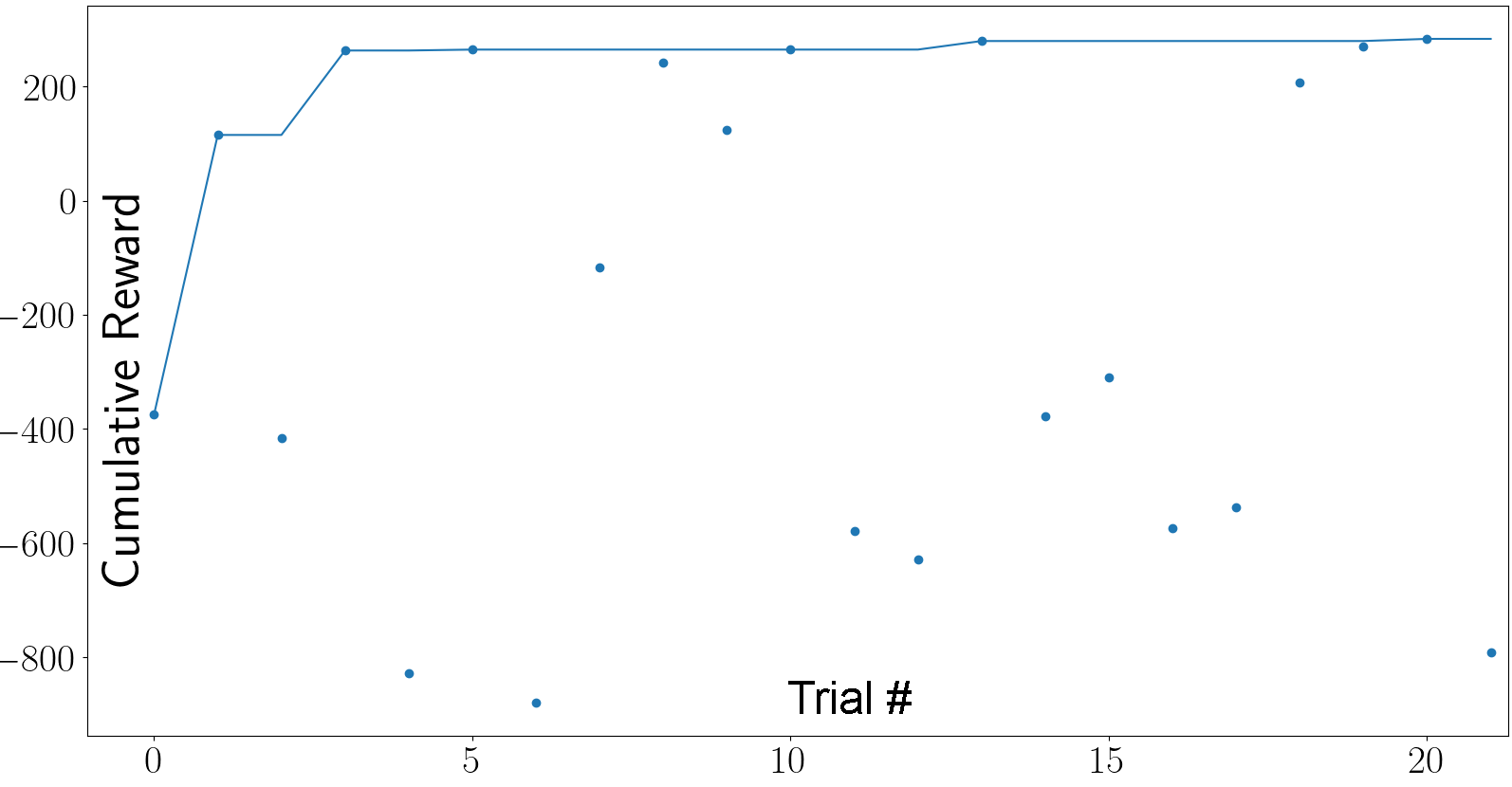}\label{fig:llA2Ctime}}
     
     \subfloat[][A2C\textsubscript{NOG+TE}]{\includegraphics[width=1.0\linewidth]{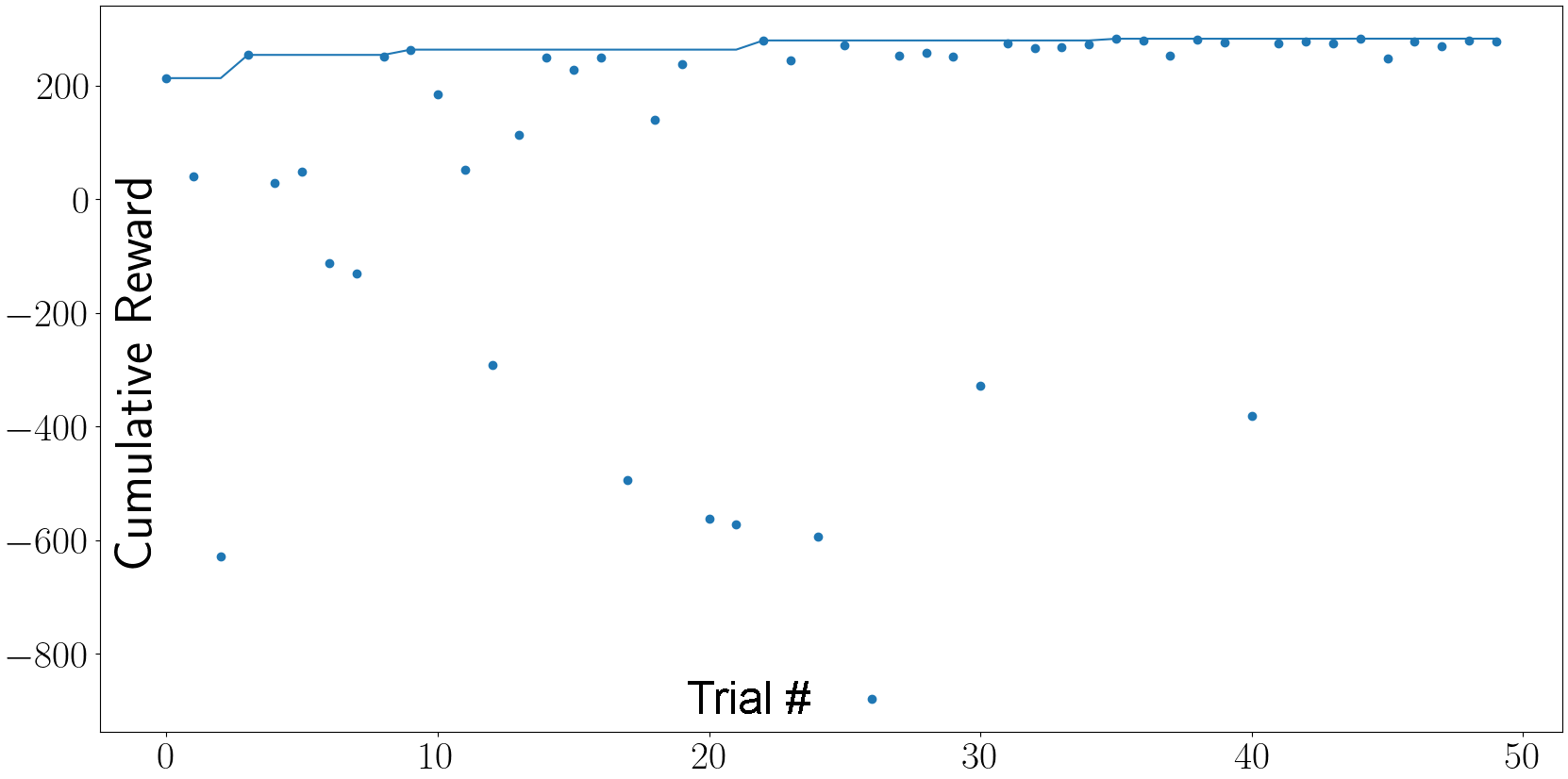}\label{fig:llA2Ctenog_time}}
     \caption{LunarLander: objective value of the hyperparameters optimization process over time, for the comparative algorithms. The solid line highlights the best values.}
     \label{fig:ll_time}
\end{figure}

Figure \ref{fig:ll_time} shows the objective value of the hyperparameters optimization process, against the number of trials sampled, for the comparative algorithm. It is worth noting that, actually, good hyperparameters can be found after just 20 trials. It is worth noting that, there is a lower number of trials in the hyperparameters optimization of A2C. Specifically, during the optimization, there are unpromising trials which are aborted during the training process by the pruning algorithm.
The figures clearly show that the A2C has been more affected by pruning with respect to the A2C\textsubscript{NOG+TE}. 

\begin{figure}[H]
     \centering
     \subfloat[][A2C]{\includegraphics[width=1.0\linewidth]{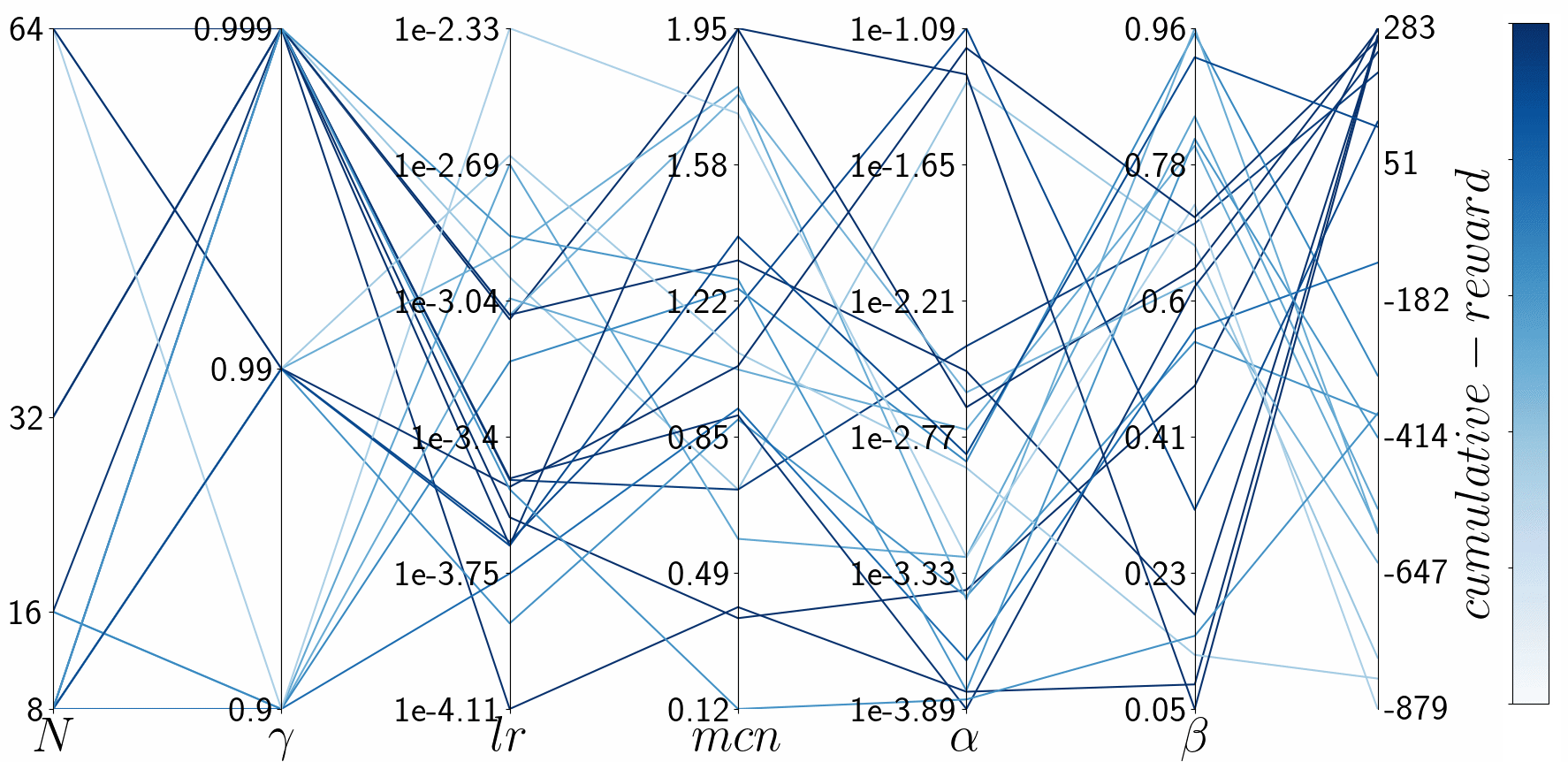}\label{fig:llA2C_parallel}}
     
     \subfloat[][A2C\textsubscript{NOG+TE}]{\includegraphics[width=1.0\linewidth]{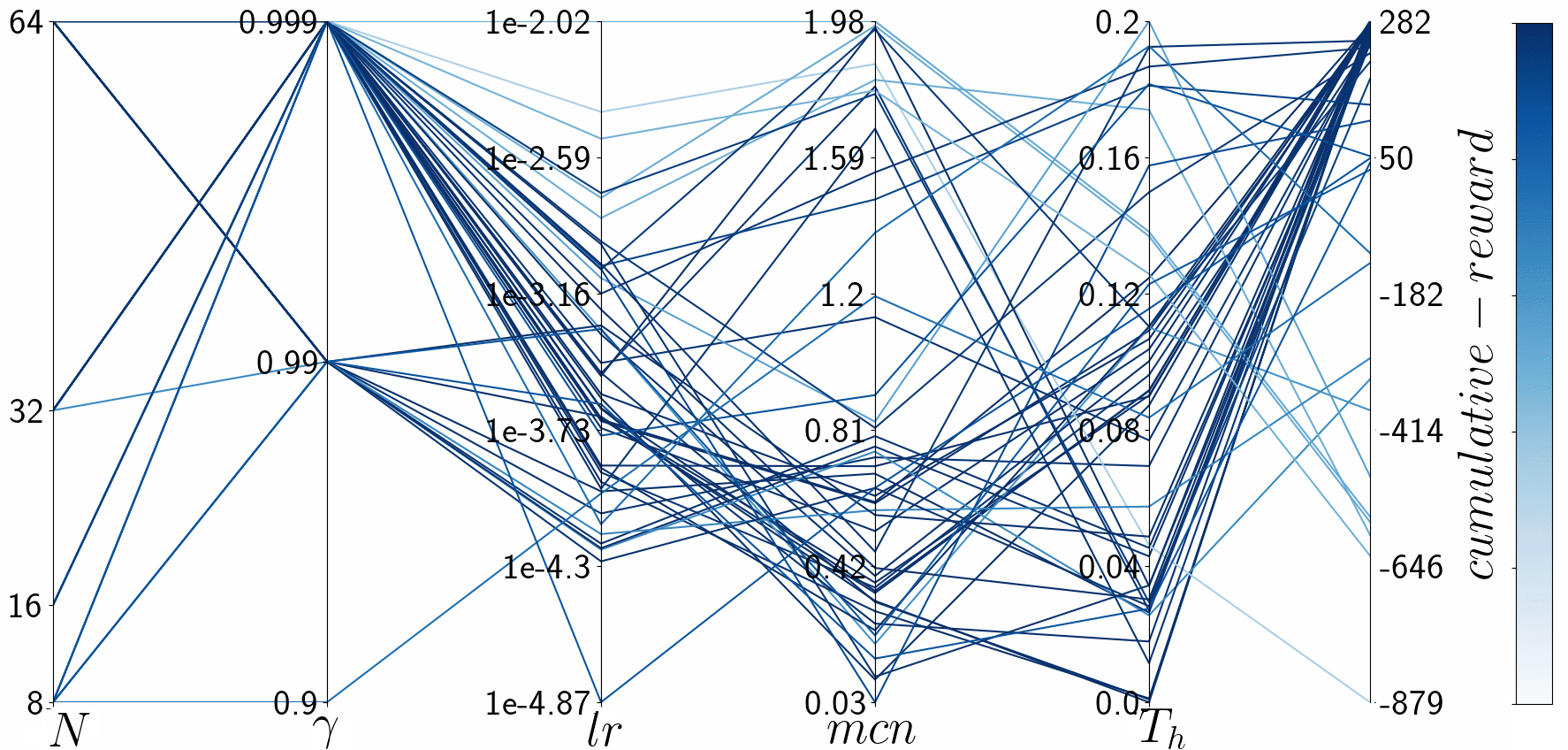}\label{fig:llA2Ctenog_parallel}}
     \caption{LunarLander: hyperparameters values optimization and related objective value, for the comparative algorithms.}
     \label{fig:ll_parallel}
\end{figure}

Figure \ref{fig:ll_parallel} represents the hyperparameters values optimization and the related objective value, for the two algorithms. Here, each line represents a trial, with its hyperparameters values represented on the vertical axes. 
According to the colorbar, the blue level of the line allows to distinguish the best solutions. In particular, in Figure \ref{fig:llA2Ctenog_parallel} it can be observed that for some hyperparameters values there is a higher density of good trials.

Table \ref{tab:ll_best_hp} shows the best hyperparameters value for each considered algorithm.

\begin{table}[!ht]
    \centering
  \begin{tabular}{||c c c ||} 
  \hline
  parameter & A2C & A2C\textsubscript{NOG+TE} \\ [0.5ex] 
  \hline\hline
  $\gamma$      & 0.999 & 0.999     \\
  \hline
  $N$           & 64        & 64        \\
  \hline
  lr            & 0.0002473 &  0.0002292 \\
  \hline
  max-clip-norm & 0.3668     & 0.3462    \\
  \hline
  $\alpha$      & 0.0003978  & \\
  \hline
  $\beta$       & 0.4832 &  \\
  \hline
  $T_h$         &  & 0.0917 \\ 
  \hline
 \end{tabular}
  \caption{LunarLander: best hyperparameters found for each algorithm.}
  \label{tab:ll_best_hp}
\end{table}

After setting the best hyperparameters for each algorithm, the training process has been carried out 10 times for each algorithm.

\begin{figure}[H]
  \centering
  \includegraphics[width=.8\linewidth]{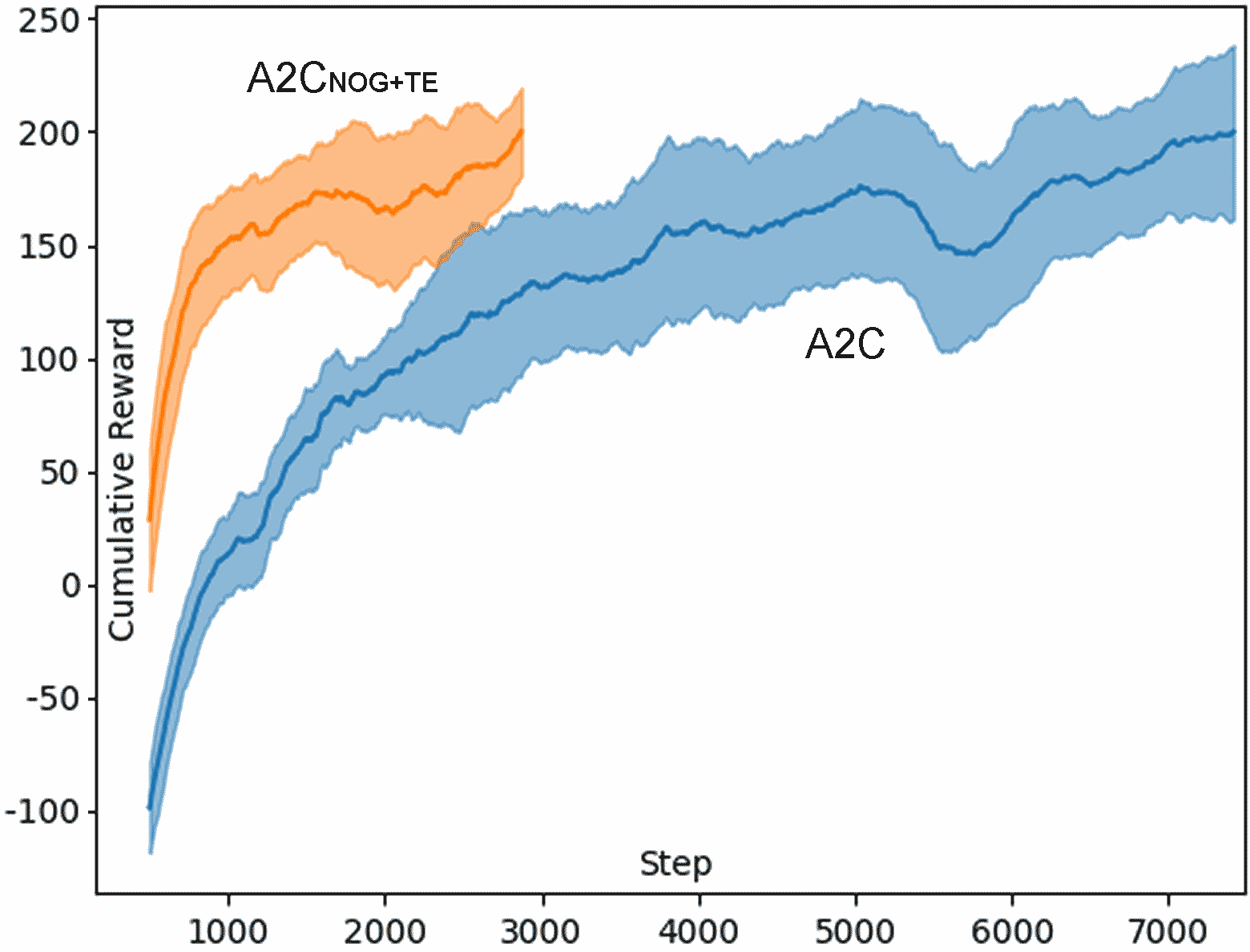}
  \caption{LunarLander: reward versus training step, for both algorithms.}
  \label{fig:lunarlander_reward_vs_training}
\end{figure}

Figure \ref{fig:lunarlander_reward_vs_training} shows the episode reward versus the training step for each algorithm, with its $95\%$ confidence interval. Precisely, the steps to solve the problem via the proposed  A2C\textsubscript{NOG+TE} algorithm and via the classical A2C are
$2045 \pm 446$ and $6265 \pm 2615$, respectively. It is apparent that the proposed approach sensibly improves the time efficiency of the A2C, up to more than 3.06x of average speedup.

\newpage

\subsection{Results summary}

Table \ref{tab:all_effic_time} summarizes
the $95\%$ confidence intervals of the training steps needed to solve the three considered environments, via A2C and A2C\textsubscript{NOG+TE}, and the speedups with respect to A2C. The effectiveness of the proposed A2C\textsubscript{NOG+TE} is apparent for increasing environment complexity (i.e., state space size).

\begin{table}[!ht]
    \centering
  \begin{tabular}{||c c c c c||} 
  \hline
  Environment & \makecell{State space \\ size} & A2C & A2C\textsubscript{NOG+TE} & \makecell{Average \\ Speedup} \\ [0.5ex] 
  \hline\hline
  EnergyMountainCar & 2 & $2702 \pm 433$ & $2511 \pm 378$ & $1.08$x \\
  \hline 
  CartPole & 4 & $999 \pm 108$ & $848 \pm 197$ & $1.18$x \\
  \hline 
LunarLander & 8 & $6265 \pm 2615$ & $2045 \pm 446$  & $3.06$x \\
  \hline 
 \end{tabular}
  \caption{Confidence intervals of the steps to solve some benchmark environments, via A2C and A2C\textsubscript{NOG+TE} algorithms.}
  \label{tab:all_effic_time}
\end{table}
The A2C\textsubscript{NOG+TE} algorithm has been developed, tested and publicly released on the Github platform \cite{a2ctenogrepo}, to foster its application on various research environments.

\section{Conclusions}

In the Advantage Actor Critic (A2C) algorithm, two issues of the scalarization of the multi-objective optimization problem are discussed and addressed. Specifically, an approach to avoid gradient overlapping (NOG) and to control the entropy (TE) of the action distribution is formally designed.
The proposed variant, called A2C\textsubscript{NOG+TE}, and the classical A2C, are experimented, after performing the hyperparameters optimization.

 The proposed techniques are designed to be used on all the reinforcement learning algorithms derived from A2C that share the same loss function components. Although the preliminary experiments look promising, more research is needed to both investigate the performance improvements on different environments and on different Advantage based algorithms.

\section*{Acknowledgements}
This research was partially carried out in the framework of the following projects: (i) PRA 2018\_81 project entitled “Wearable sensor systems:
personalized analysis and data security in healthcare”
funded by the University of Pisa; (ii) CrossLab project (Departments of Excellence), funded by the Italian Ministry of Education and Research (MIUR); (iii) “KiFoot: Sensorized footwear for gait analysis” project, co-funded by the Tuscany Region (Italy) under the PAR FAS 2007-2013 fund and the FAR fund of the Ministry of Education, University and Research (MIUR).

\bibliography{bibliography}
\bibliographystyle{ieeetr}

\section*{Authors}
\begin{description}
\item[Federico A. Galatolo] is a PhD student in Information Engineering at the Department of Information Engineering of the University of Pisa (Italy). His research is focused on Computational Stigmergy, Deep Learning and Reinforcement Learning.

\item[Mario G.C.A. Cimino] is an associate professor at the Department of Information Engineering of the University of Pisa (Italy). His research lies in the areas of Information Systems and Artificial intelligence. He is (co-) author of about 70 scientific publications.

\item[Gigliola Vaglini] is full professor of Computer Engineering at ``Dipartimento di Ingegneria della Informazione'' of the University of Pisa. The main fields of her research activity are the specification and verification of concurrent and distributed systems, and the use of machine learning techniques for detecting malware in mobile systems and anomalies of industrial systems.
\end{description}
\end{document}